\documentclass[a4paper,fleqn]{cas-dc}

\usepackage[numbers]{natbib}
\usepackage{appendix}
\usepackage{chngcntr}  
\usepackage{float}
\restylefloat{figure}  
\usepackage{amsmath,amsfonts,amssymb,mathrsfs}
\usepackage{xcolor}
\usepackage{rotating}
\usepackage{booktabs} 
\usepackage{threeparttable}
\usepackage{subcaption}
\usepackage[normalem]{ulem}
\usepackage{graphicx}

\usepackage{balance}

\usepackage{tikz}
\usetikzlibrary{arrows.meta,positioning,fit,calc,decorations.pathreplacing}
\usepackage{adjustbox}

\usepackage{algorithm}
\usepackage{algpseudocode}
\usepackage{amsmath,amssymb}
\usepackage{multirow}
\usepackage{xurl}
\usepackage{amssymb}

\usepackage{textcomp}

\hypersetup{
  hidelinks,
  breaklinks=true
}

\expandafter\def\csname __first_footerline:\endcsname{%
  \begingroup
  \small\sffamily
  \csname __short_authors:\endcsname
  \endgroup
}

\definecolor{accessblue}{RGB}{0,105,154}

\makeatletter
\@ifundefined{xfigwd}{\newlength{\xfigwd}}{}
\makeatother

\def\BibTeX{{\rm B\kern-.05em{\sc i\kern-.025em b}\kern-.08em
    T\kern-.1667em\lower.7ex\hbox{E}\kern-.125emX}}

\def\tsc#1{\csdef{#1}{\textsc{\lowercase{#1}}\xspace}}
\tsc{WGM}
\tsc{QE}
\tsc{EP}
\tsc{PMS}
\tsc{BEC}
\tsc{DE}

\begin{document}
\def\floatpagepagefraction{1}
\def\textpagefraction{.001}
\shortauthors{M.Alavi et al.}

\title [mode = title]{Generating Bearing Vibration Signals at User-Specified Fault Probabilities Using PR-GAN and Counterfactual Methods}
\shorttitle{Generating Bearing Vibration Signals at User-Specified Fault Probabilities Using PR-GAN and Counterfactual Methods}

\author[1]{Seyed Mohammadreza Alavi}[type=editor,
   auid=,bioid=,
   orcid=]
\ead{}
\credit{We add the roles once we finish the paper}

\author[2]{Ardeshir Shojaeinasab}[type=editor,
   auid=,bioid=,
   orcid=]
\ead{}
\credit{We add the roles once we finish the paper}

\author[3]{Reza Jalayer}[type=editor,
   auid=,bioid=,
   orcid=]
\ead{}
\cormark[1]
\credit{We add the roles once we finish the paper}

\author[4]{Masoud Jalayer}[type=editor,
   auid=,bioid=,
   orcid=]
\ead{}
\credit{We add the roles once we finish the paper}

\author[2]{Behnam Bahrak}[type=editor,
   auid=,bioid=,
   orcid=]
\ead{}
\credit{We add the roles once we finish the paper}

\address[1]{Department of Electrical and Computer Engineering, University of Tehran, Tehran, Iran}
\address[2]{Department of Electrical and Computer Engineering, University of Victoria, Victoria BC, Canada}
\address[3]{FAST-Lab, Faculty of Engineering and Natural Sciences, Tampere University, Korkeakoulunkatu 7, 33720 Tampere, Finland}
\address[4]{Department of Information and Communications Engineering, Aalto University, 02150 Espoo, Finland}

\cortext[cor1]{Corresponding author}

\begin{abstract}
In bearing vibration datasets, most samples receive predicted fault probabilities close to 0 or 1, while samples with intermediate (gray-zone) probabilities are rare. Such borderline samples are practically important because they reflect conditions in which maintenance decisions often require additional inspection or a more conservative response, and they are also useful for studying decision boundaries. To address this scarcity, this paper proposes and compares two approaches that generate vibration signals whose predicted fault probability matches a specific target $p^{*}\in\{0.25,0.50,0.75\}$. We use the average output of a heterogeneous ensemble classifier whose members use different architectures and random initializations as a fixed, gradient-accessible probability oracle. The first approach is a training-based method, Probability-Regularized Generative Adversarial Network (PR-GAN), which extends Wasserstein Generative Adversarial Network with Gradient Penalty (WGAN-GP) and edits a real signal through a residual generator while pushing the classifier output toward the target probability. The second approach instantiates a Wachter-style counterfactual (CF) procedure, training-free and per-sample, that directly optimizes each input signal to reach the target probability while remaining close to the source signal. We evaluate both methods on the Case Western Reserve University (CWRU) and Paderborn bearing datasets using (i) mean absolute target-probability error, (ii) total variation in the time domain, and (iii) log power spectral density (log-PSD) differences in the frequency domain. Across all settings, CF reaches the target with a mean absolute probability error of $0.005$--$0.008$ and a within-tolerance success rate of $1.000$ on retained samples, whereas PR-GAN's mean error is $0.046$--$0.059$ with success rates between $0.501$ and $0.680$. CF therefore steers the probability more reliably and requires smaller average $L_1$ changes, whereas PR-GAN has a lower reported runtime in most settings.
\end{abstract}

\maketitle

\section{Introduction}
\label{sec:introduction}
Rolling-element bearings are core components in many rotating machines, and bearing faults are a common cause of unexpected failures and costly downtime~\cite{zhang2020deep,jalayer2024evaluating}. In recent years, deep learning models have improved data-driven bearing fault diagnosis by learning features from vibration signals~\cite{wen2018cnn,shao2017autoencoder}.

Despite strong classification performance, samples with intermediate predicted fault probabilities are rare in practice: most real signals are confidently scored as either healthy or faulty under a trained classifier, and the classifier's posterior is therefore concentrated near 0 and 1. The scarcity is a property of the classifier's output distribution rather than of the underlying signals themselves, but its consequence is the same. Few examples are available in the boundary region (i.e., when $P_{\mathrm{fault}}({x})$ is close to 0.5), which limits the systematic study of model behavior in ambiguous operating conditions.

Prior generative augmentation methods (e.g., Variational Autoencoders (VAEs) and Generative Adversarial Network (GAN) variants) can increase the number of fault samples, but they typically generate class-conditional signals (normal vs.\ fault) and do not explicitly control a trained classifier’s posterior probability~\cite{mao2019gan,tang2021wgan,cheng2024cvae}. In a complementary direction, counterfactual methods modify an input to achieve a desired model prediction, but ensuring that the modified input remains a valid (realistic) data point can be challenging~\cite{wachter2018gdpr}.

To address the lack of intermediate-probability samples, we study \emph{probability-targeted synthesis} for bearing vibration signals. Given a real signal window $x$ and a target fault probability $p^{*}\in\{0.25,0.50,0.75\}$, our goal is to generate a realistic signal $\hat{x}$ such that a fixed classifier assigns $P_{\mathrm{fault}}(\hat{x})\approx p^{*}$. We use an ensemble CNN as a fixed, gradient-accessible probability oracle and investigate two generation approaches: (i) a Probability-Regularized Generative Adversarial Network (PR-GAN) that extends Wasserstein Generative Adversarial Network with Gradient Penalty (WGAN-GP)~\cite{gulrajani2017improved} with a probability-alignment objective, and (ii) a direct optimization-based counterfactual method that updates the signal with small steps to match the target probability while remaining close to the original sample.

We consider two generation paradigms. PR-GAN is a \emph{training-based} approach: it requires learning a generator via adversarial training before synthesis. In contrast, the counterfactual method is \emph{training-free}: it produces each sample by directly optimizing the input signal at inference time using the fixed classifier, without training an additional generative model.

We evaluate both methods on the CWRU and Paderborn bearing datasets using a binary classification setting: normal and fault. Our evaluation focuses on (1) probability-steering accuracy via the mean absolute target-probability error, and (2) signal realism via time-domain total variation (TV) and frequency-domain log power spectral density (log-PSD) differences.

Our main contributions are:
\begin{itemize}
    \item We formulate probability-targeted generation for bearing vibration signals and study target probabilities $p^{*}\in\{0.25,0.50,0.75\}$ using an ensemble CNN as a probability oracle.
    \item We develop and compare two complementary approaches for probability control: PR-GAN (classifier-in-the-loop adversarial training) and CF (optimization-based counterfactual synthesis).
    \item We provide a quantitative evaluation on CWRU and Paderborn that analyzes probability steering and signal realism in both time and frequency domains.
\end{itemize}

Our aim is to use the methods to generate new samples that (1) meet the target probability and (2) remain close to their original source samples.

\paragraph{Positioning of the contribution.} The CF method we use is a Wachter-style counterfactual~\cite{wachter2018gdpr}: a per-sample optimization of a weighted sum of a classification loss and a proximity term to the source. The novelty of the present work is not the form of this objective but its application. We apply it to bearing vibration signals rather than tabular or image inputs, and we use a real-valued probability target $p^{*}\in\{0.25,0.50,0.75\}$ rather than the binary class flip that is typical in counterfactual explanation. PR-GAN combines several existing ingredients: the WGAN-GP critic and gradient penalty~\cite{gulrajani2017improved}, classifier-in-the-loop posterior alignment as in GAP~\cite{Chhabra-GAP}, and the residual editing idea found in earlier classifier-guided generation~\cite{nguyen2017ppgn}. The aspect specific to PR-GAN is the use of a continuous probability-alignment term (rather than a class-score term) inside this combination, on vibration windows. We therefore present both methods as concrete instantiations of well-known generative templates on bearing vibration signals; the empirical question we ask is which template is more faithful for probability-targeted generation in this domain.

\section{Related Work}
\label{sec:related_work}
We organize prior work into six areas based on their main focus, which frame our study: (i) deep learning for bearing fault diagnosis; (ii) generating vibration signals with GANs; (iii) borderline oversampling for class imbalance; (iv) calibration and uncertainty; (v) counterfactual explanations for time-series fault diagnosis; and (vi) classifier-guided, probability-controlled generation. Across these areas, many methods generate label-conditional samples or aim only to cross a decision threshold, while few control the model’s predicted probability to a user-specified value. Our framework addresses this need.

\subsection{Deep Learning for Bearing Fault Diagnosis}
Deep learning is widely used for vibration-based bearing diagnosis. Surveys report that convolutional neural networks (CNNs) and autoencoders achieve high accuracy on laboratory datasets such as CWRU by learning features directly from data~\cite{zhang2020deep}. Supervised CNNs often operate on time-domain signals or on 2-D representations and outperform traditional baselines~\cite{wen2018cnn}. Autoencoder-based representations further reduce reliance on hand-crafted features and can improve robustness to noise and operating variation~\cite{shao2017autoencoder}. A practical challenge is class imbalance; Qian and Li~\cite{qian2020imbalance} proposed a class-imbalance-robust network for bearing fault diagnosis that uses BSF for feature extraction and BSOF for classification. Their method addresses class imbalance by balancing discriminative feature extraction across classes and reducing weight-updating imbalance. They also group class-imbalance methods into data-level, classifier-level, and algorithm-level approaches.

\subsection{Generating Vibration Signals with GANs}
Prior work uses GANs to augment minority fault classes and rebalance training data on bearing benchmarks such as CWRU~\cite{mao2019gan,luo2021cdcgans,tang2021wgan}. Pipelines often convert vibration signals to spectra or images and train conditional or Wasserstein variants; for example, WGAN-GP is used for stable augmentation in rolling-bearing tasks~\cite{tang2021wgan,ruan2023improvement}. To handle operating-condition changes, N2FGAN frames generation as normal-to-fault translation to produce fault data for unseen speeds or loads~\cite{ahang2022synthesizing}. Hybrid designs such as CVAE–SKEGAN couple variational and adversarial objectives and report gains under imbalance and limited labels~\cite{cheng2024cvae}. A recent review shows that GAN-based bearing-fault generation has mainly been used for data augmentation for small-sample and imbalanced conditions, while conditional variants generate samples of specified fault classes and Wasserstein-based variants improve training stability~\cite{ruan2023improvement}.

\subsection{Borderline Oversampling for Class Imbalance}
Beyond label-conditional augmentation, several methods synthesize new minority samples \emph{near the decision boundary} to address class imbalance. OBGAN generates minority data close to the borderline using a GAN-based oversampling scheme and reports improved performance on imbalanced classification tasks~\cite{jo2022obgan}. ESLEBS extends Borderline-SMOTE with a \emph{safe-level} mechanism that places synthetic samples in safer minority regions near the boundary, improving minority recognition while reducing risky placements~\cite{Chen2023-ESLEBS}. Adaptive SV-Borderline SMOTE-SVM uses minority-class support vectors as boundary references and synthesizes new samples around these points to enhance classification on imbalanced datasets~\cite{Guo2024-SVBorderlineSMOTE}. Borderline oversampling and our setting are related but distinct: those methods synthesize samples near a binary decision boundary in order to rebalance training data, whereas our setting synthesizes samples at a user-specified continuous posterior value $p^{*}$ regardless of class balance. The two settings coincide only at $p^{*}=0.5$ for a balanced binary classifier; otherwise they target different points in posterior space.

\subsection{Calibration and Uncertainty}
Modern deep networks are often miscalibrated: their predicted probabilities do not match observed frequencies. Standard tools to assess this include reliability diagrams and expected calibration error (ECE), and a simple post hoc method-temperature scaling can reduce ECE without changing accuracy~\cite{Guo2017-Calibration}.
Uncertainty is also important for safe deployment. Epistemic (model) uncertainty can be estimated with Monte Carlo dropout by keeping dropout active at test time and averaging predictions, which provides a practical Bayesian approximation~\cite{gal2016dropout}. In rotating machinery fault diagnosis, recent work evaluates epistemic and aleatoric (data) uncertainty on CWRU and compares approaches such as MC dropout, Bayesian neural networks, and deep ensembles; the study reports that ensembles often provide strong out-of-distribution detection and robust performance under noise~\cite{jalayer2024evaluating}.
More broadly, probability-targeted generation may support uncertainty-aware fault diagnosis by adding training samples from a wider range of predicted fault probabilities.

\subsection{Counterfactual Explanations for Time-Series Fault Diagnosis}
Counterfactual explanations aim to modify an input with minimal change so that a fixed classifier produces a desired prediction (class or score)~\cite{wachter2018gdpr}. For time series, recent methods add locality and plausibility constraints to keep edits realistic; examples include instance-based counterfactuals that perturb contiguous subsequences~\cite{delaney2021instance} and guided, locally constrained counterfactuals that impose temporal constraints while achieving a target class~\cite{wang2024glacier}.
In machinery fault diagnosis, counterfactuals have been used to improve generalization with few labels by augmenting contrastive learning~\cite{yliu2024fewshot}, to support open-set recognition under speed transients via representation-aligned counterfactual objectives~\cite{liu2025aligned}, and to enable generalized zero-shot recognition of compound faults through counterfactual inference~\cite{xu2025zero}. Counterfactual reasoning has also been applied in process industries (e.g., batch fermentation) for fault diagnosis~\cite{liu2024batch}. In addition, causality-aware counterfactual data generation integrates physical priors with generative models to produce class-consistent samples for electromechanical systems~\cite{wang2025cfdata}.

\subsection{Classifier-Guided, Probability-Controlled Generation}
Classifier-guided generation uses a trained classifier during synthesis to steer samples toward desired predictions. Plug-and-Play Generative Networks (PPGN) exemplify this idea by updating an input or latent code to increase a classifier’s class score while staying near the data manifold through a learned prior~\cite{nguyen2017ppgn}. Autoencoder theory supports such manifold constraints: denoising and related regularized autoencoders estimate the score of the data distribution and thus encourage edits to remain on-manifold~\cite{alain2014dae,Bengio2013-GDAE}.
A complementary direction aligns classifier \emph{posterior} outputs directly. Generative Alignment of Posterior Probabilities (GAP) matches target-domain classifier posteriors to those of a source model, providing a classifier-in-the-loop objective in label space~\cite{Chhabra-GAP}. For adversarial components, standard practices improve stability and diversity: WGAN-GP enforces a 1-Lipschitz critic via a gradient penalty, and minibatch discrimination helps avoid mode collapse~\cite{gulrajani2017improved,salimans2016improved}. During iterative editing, total variation regularization suppresses spurious high-frequency oscillations while preserving salient structure~\cite{Getreuer2012-ROF-SB}.

\section{Methods}
\label{sec:methods}
This section describes our method for generating samples with a target classifier probability. We use an ensemble CNN as the probability oracle. In this paper, \(f(x)\) denotes the fault probability predicted by the model for an input signal \(x\). We then generate new samples with $0.25$, $0.5$, and $0.75$ fault probabilities using two approaches: PR-GAN and an optimization-based counterfactual method. We generated 500 samples for each experimental setting (dataset, and target probability) per method, and all experiments were run on A100 GPU. The implementation is available on GitHub repository\footnote{\href{https://github.com/mralavi20/Generating-Bearing-Vibration-Signals-at-User-Specified-Fault-Probabilities}{\nolinkurl{https://github.com/mralavi20/Generating-Bearing-Vibration-Signals-at-User-Specified-Fault-Probabilities}}}.

\subsection{Training-based and training-free approaches}
We distinguish two families of generative procedures that we will use throughout. \emph{Training-based} methods learn a generator on a training set in an offline phase; once trained, the generator is frozen and produces each sample with a single forward pass at inference. Training is the bottleneck and any change in objective or hyperparameters typically requires retraining. \emph{Training-free} methods skip the offline phase and instead optimize an output for each input at generation time, using a fixed model only for guidance. Generation is therefore the bottleneck, settings can be changed per sample, and outcomes depend on the input and the optimization configuration. PR-GAN belongs to the first family, the counterfactual method to the second.

\subsection{Using a fixed, gradient-accessible guidance model}
\label{sec:reliable_guidance}
\begin{figure*}[t]
  \centering
  \includegraphics[width=0.6\linewidth]{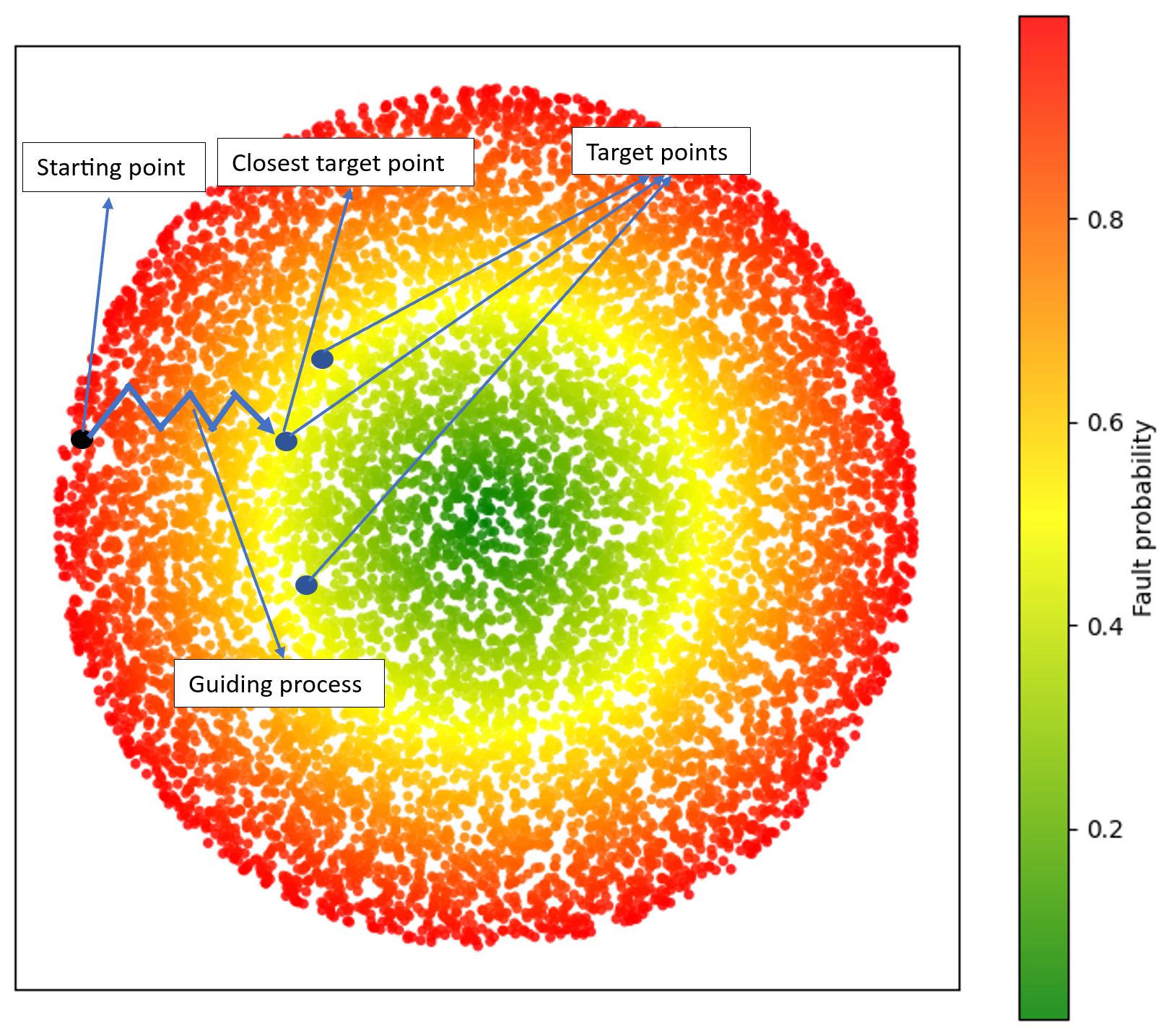}
  \caption{Schematic illustration of the proposed search process in the latent space.}
  \label{fig:latent_space_search}
\end{figure*}

\begin{figure}[t]
\centering
\begin{adjustbox}{width=0.6\columnwidth}
\begin{tikzpicture}[
  font=\small,
  node distance=7mm and 10mm,
  box/.style={draw, rounded corners, thick, align=center, inner sep=4pt, fill=#1!7},
  arrow/.style={-{Latex[length=2.2mm]}, thick},
  lab/.style={font=\scriptsize, inner sep=1pt}
]

\node[box=gray, minimum width=48mm, minimum height=20mm] (gen)
{Generating process};

\draw[arrow] ([xshift=-14mm]gen.west) -- (gen.west)
node[midway, above] {input};

\draw[arrow] (gen.east) -- ([xshift=14mm]gen.east)
node[midway, above] {output};

\node[box=green, above=14mm of gen, minimum width=40mm, minimum height=18mm] (oracle)
{Guidance model};

\draw[arrow] ($(oracle.south)+(-10mm,0)$) -- ($(gen.north)+(-10mm,0)$);

\draw[arrow] ($(gen.north)+(10mm,0)$) -- ($(oracle.south)+(10mm,0)$);

\end{tikzpicture}
\end{adjustbox}
\caption{A fixed, gradient-accessible guidance model directs the generating process from input to output.}
\label{fig:reliable_model_guidance}
\end{figure}
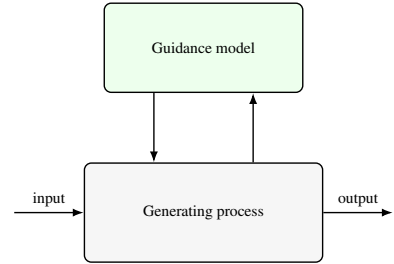

Both approaches share a fixed external classifier as a guidance model. By \emph{fixed} we mean that the classifier is trained once on labeled data and then frozen during generation; its weights are never updated by either PR-GAN or the counterfactual procedure. By \emph{gradient-accessible} we mean that, although we treat the classifier's internal architecture and decision logic as opaque, we do read its gradients with respect to the input during optimization. The classifier therefore plays the same functional role in both approaches: it receives a candidate signal and returns a fault probability, and that probability (together with its gradient) drives the generator or the counterfactual updates toward the target $p^{*}$ (Figure~\ref{fig:reliable_model_guidance}). Fig.~\ref{fig:latent_space_search} illustrates an example of the probability-guided search process in a schematic two-dimensional latent space.

We assume throughout this paper that the classifier's predicted probabilities are sufficiently informative to guide generation. We do not establish that they are well calibrated, and we do not measure expected calibration error or reliability diagrams on the ensemble; this is a deliberate scoping choice that lets us isolate the behavior of the two generation procedures from any choices about the guidance model. The implication, which we revisit in the discussion, is that a generated signal with $f(\hat{x})\!\approx\!0.5$ should be read as ``the classifier is uncertain about $\hat{x}$'' rather than as a calibrated 50\% fault likelihood.

\subsection{Continuity of the signal space and the probability oracle}
In real machines, vibration signals can be treated as reflecting an underlying continuous-time vibration response because they come from physical motion governed by inertia, stiffness, and damping. Therefore, the underlying response tends to change gradually over time, which makes the local continuity assumption reasonable~\cite{BlakeBasicVibrationTheory}.

The probability oracle can also be treated as a continuous function of the input \(x\), because the CNN inference pipeline is composed of continuous operations, and the ensemble output is obtained by averaging the member probabilities~\cite{Gouk2020LipschitzContinuity,Ju2017EnsembleCNN}.

\subsection{Existence of a signal with target predicted fault probability}

In this part, we provide a simple existence argument that at least one signal exists with target predicted fault probability $p^{*}$.

Assume:
\begin{itemize}
    \item $\Omega$, the space of admissible signals, is a continuous space and there is a continuous path between any two signals without leaving the space and it is path-connected.
    \item The model’s probability function $f(x)$ is continuous over $\Omega$.
    \item Two signals $x_{1}, x_{2} \in \Omega$ exist such that:
    \begin{equation}
        f(x_{1}) < p^{*} < f(x_{2})
    \end{equation}
\end{itemize}

Then:
\begin{equation}
    \text{There exists} \;\; x^{*} \in \Omega \;\; \text{such that} \;\; f(x^{*}) = p^{*}.
\end{equation}

\paragraph{Existence argument}
By path-connectivity, there exists a continuous path $\gamma : [0,1] \rightarrow \Omega$ with $\gamma(0)=x_{1}$ and $\gamma(1)=x_{2}$. Define:
\begin{equation}
    g(t) = f(\gamma(t)).
\end{equation}

Since $\gamma$ and $f$ are continuous, $g$ is continuous too. Also:
\begin{equation}
    g(0) = f(x_{1}) < p^{*}, \qquad g(1) = f(x_{2}) > p^{*}.
\end{equation}

By the Intermediate Value Theorem, there exists $t^{*} \in (0,1)$ such that $g(t^{*}) = p^{*}$. Let $x^{*} = \gamma(t^{*})$. Then $x^{*} \in \Omega$ and $f(x^{*}) = p^{*}$.

This argument does not guarantee that a particular algorithm will always find such a signal. It also addresses only the easier of two questions: it shows that some signal in the unconstrained space $\Omega$ achieves $f(x^{*})=p^{*}$, but it does not show that any such signal lies on the bearing-vibration data manifold. Whether a target-probability signal that is also realistic (i.e., that resembles a plausible vibration window for the operating condition) exists is left to the empirical evaluation; the time-domain and frequency-domain similarity metrics in Section~\ref{sec:results_discussion} are intended as proxies for that question.

\subsection{Datasets}
\begin{figure}[t]
  \centering
  \includegraphics[width=\linewidth]{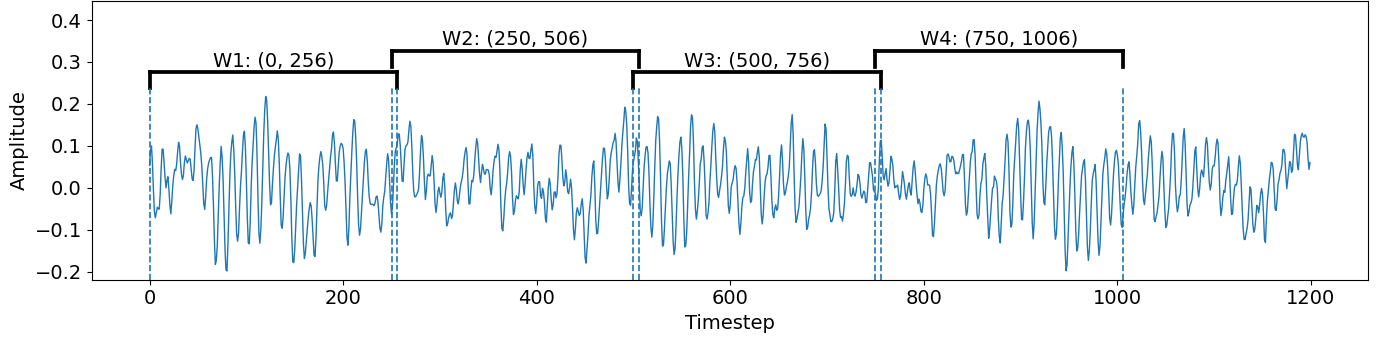}
  \caption{Sliding-window cropping with length=256 and stride=250 on a raw signal.}
  \label{fig:window_brackets}
\end{figure}

We use a subset of the Case Western Reserve University (CWRU)~\cite{CWRU_BearingDataCenter} and Paderborn~\cite{PaderbornBDC_DataSetsDownload} bearing datasets. We split the data into 70\% training, 15\% validation, and 15\% test sets. For each dataset, we form a binary classification setting with class 0 as the normal (healthy) condition and class 1 as the fault condition. In this binary setting, all fault types are grouped into the single class ``fault''. For the fault class, we attempted to maintain a balanced distribution across different fault types.
For CWRU, we have three fault categories: inner race (IR), outer race (OR), and ball/rolling element (B) faults. For Paderborn, we have four types: real and artificial inner ring (IR) faults, real and artificial outer ring (OR) faults. We segment each raw time-series into overlapping windows of length 256 samples and use a sliding window with a stride of 250 samples (Figure~\ref{fig:window_brackets}). For CWRU, the drive-end accelerometer signal and fan-end accelerometer signal were used as two input channels, whereas for Paderborn, the selected bearing-housing vibration signal was used as a single input channel.

\subsubsection{CWRU dataset}
\begin{figure}[t]
  \centering
  \includegraphics[width=0.6\linewidth]{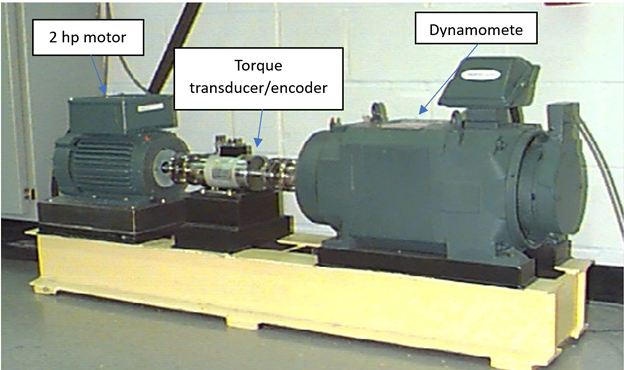}
  \caption{Experimental setup of the CWRU bearing test rig (adapted from~\cite{CWRU_BearingDataCenter} and annotated).}
  \label{fig:cwru-dataset}
\end{figure}

The Case Western Reserve University (CWRU) bearing dataset is a widely used open-source benchmark available through the CWRU website for normal and faulty bearing data~\cite{Neupane2020BearingFaultCWRUReview}. This dataset provides vibration recordings collected on a laboratory test for normal bearings, single point drive-end (DE), and fan-end (FE) defects with sample frequency of 12 kHz and 48 kHz. Fault labels are commonly defined by defect location, inner-race (IR), outer-race (OR), and ball/rolling element (B)~\cite{Neupane2020BearingFaultCWRUReview}. Figure~\ref{fig:cwru-dataset} shows experimental setup of the CWRU bearing test rig (adapted from~\cite{CWRU_BearingDataCenter} and annotated).

\subsubsection{Paderborn dataset}
\begin{figure}[t]
  \centering
  \includegraphics[width=0.8\linewidth]{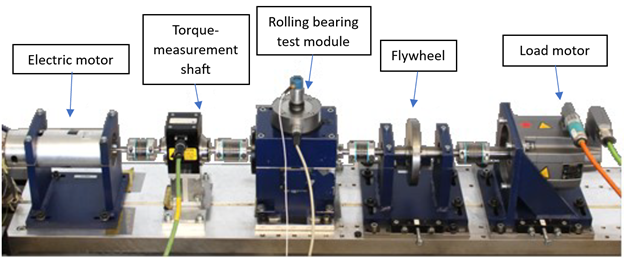}
  \caption{Experimental setup of the Paderborn bearing test rig (adapted from Fig.~4 in~\cite{lessmeier2016kat_benchmark} and annotated).}
  \label{fig:paderborn-dataset}
\end{figure}

The Paderborn bearing dataset provides synchronously recorded motor current signals (MCS) together with reference vibration acceleration, collected on an electromechanical drive-system test rig and digitized synchronously at 64 kHz~\cite{lessmeier2016kat_benchmark}. The dataset contains 32 bearing experiments (6 healthy, 12 artificially damaged, and 14 real-damage bearings from accelerated lifetime tests)~\cite{lessmeier2016kat_benchmark}. The dataset distinguishes artificial damage from real damage and reports that both occur at the inner ring (IR) and outer ring (OR), enabling four location-by-origin categories: real IR, real OR, artificial IR, and artificial OR~\cite{lessmeier2016kat_benchmark}. Figure~\ref{fig:paderborn-dataset} shows experimental setup of the Paderborn bearing test rig (adapted from Fig.~4 in~\cite{lessmeier2016kat_benchmark} and annotated).

\subsection{Baseline classifier}
The predicted probabilities of classifiers are not always well calibrated or interpretable. To reduce this issue, we use an ensemble classifier and compute the average predicted probability across ensemble members. Our baseline classifier is a heterogeneous ensemble classifier. For each input window, we compute the predicted class probabilities by averaging the probability outputs of all ensemble members. This averaging reduces overconfident predictions and typically yields a more informative probability distribution. The ensemble members use different architectures and random initializations (seeds) and, as a result, follow different training trajectories. The model architecture for each member is summarized in table~\ref{tab:classifier_members}. Following the scoping choice stated in Section~\ref{sec:reliable_guidance}, we do not analyze the ensemble's internal calibration further; we report its accuracy and F1 in Section~\ref{sec:results_discussion} only to establish that the classifier is competent enough on the underlying classification task to be used as a guidance signal.

\begin{table*}[t]
\centering
\caption{Structure of the heterogeneous ensemble classifier members.}
\label{tab:classifier_members}
\small
\renewcommand{\arraystretch}{1.15}
\begin{tabular}{p{0.24\textwidth} p{0.68\textwidth}}
\hline
\textbf{Member type} & \textbf{Architecture} \\
\hline
Deep CNN & Conv1D--BN--ReLU--MP--Dropout $\rightarrow$ Conv1D--BN--ReLU--MP--Dropout $\rightarrow$ Conv1D--BN--ReLU--MP--Dropout $\rightarrow$ GAP $\rightarrow$ Dense--BN--ReLU--Dropout $\rightarrow$ Softmax \\
Shallow CNN & Conv1D--BN--ReLU $\rightarrow$ GAP $\rightarrow$ Softmax \\
ConvLSTM & Reshape $\rightarrow$ ConvLSTM1D--BN $\rightarrow$ GAP $\rightarrow$ Softmax \\
Dilated CNN & Dilated Conv1D--BN--ReLU $\rightarrow$ GAP $\rightarrow$ Softmax \\
Multi-scale CNN & Parallel Conv1D branches $\rightarrow$ Concatenation $\rightarrow$ BN--ReLU $\rightarrow$ GAP $\rightarrow$ Softmax \\
Separable CNN & SeparableConv1D--BN--ReLU $\rightarrow$ GAP $\rightarrow$ Softmax \\
GRU & GRU $\rightarrow$ GAP $\rightarrow$ BN $\rightarrow$ Softmax \\
BiLSTM & BiLSTM $\rightarrow$ GAP $\rightarrow$ BN $\rightarrow$ Softmax \\
MC-dropout CNN & Conv1D--ReLU--Dropout $\rightarrow$ GAP $\rightarrow$ Softmax $\rightarrow$ MC averaging \\
Mixup dilated CNN & Dilated Conv1D--BN--ReLU $\rightarrow$ GAP $\rightarrow$ Softmax; mixup training \\
\hline
\end{tabular}

\vspace{1mm}
\begin{minipage}{0.92\textwidth}
\footnotesize \textit{Note:} GAP denotes global average pooling.
\end{minipage}
\end{table*}

\subsection{Probability-Regularized GAN (PR-GAN)}
\label{sec:prgan}
We propose Probability-Regularized GAN (PR-GAN) as a \emph{training-based} approach to generate new samples whose predicted fault probability matches a specified target value. PR-GAN extends WGAN-GP~\cite{gulrajani2017improved} by incorporating a fixed classifier as a probability oracle and adding a probability-alignment term to the generator objective. The classifier-in-the-loop construction is closest to GAP~\cite{Chhabra-GAP}, which aligns posterior probabilities, and to PPGN~\cite{nguyen2017ppgn}, which steers samples toward classifier outputs; PR-GAN differs from these in two respects relevant to our setting: (i) it aligns to a continuous probability target $p^{*}$ rather than to a class-score peak, and (ii) it operates on vibration windows through a residual edit of an existing real signal rather than synthesizing from a latent code. We run PR-GAN for 300 epochs.

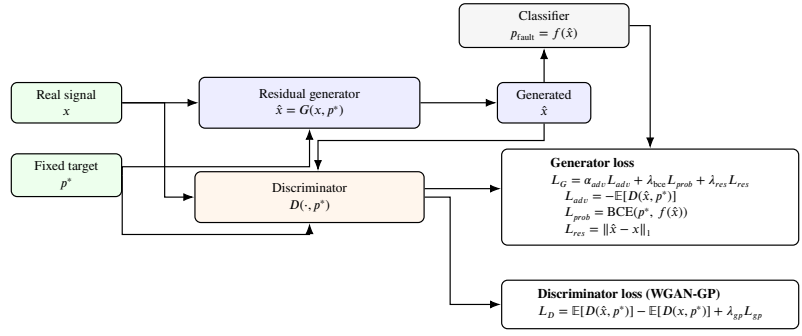
\begin{figure*}[t]
\centering
\begin{adjustbox}{width=0.60\textwidth}
\begin{tikzpicture}[
  font=\small,
  node distance=10mm and 16mm,
  box/.style={draw, rounded corners, thick, align=center, inner sep=5pt, fill=#1!7},
  loss/.style={draw, rounded corners, thick, align=left, inner sep=6pt, fill=#1!5},
  arrow/.style={-{Latex[length=2.6mm]}, thick},
  lab/.style={font=\scriptsize, inner sep=1pt}
]

\node[box=green, minimum width=26mm] (x) {Real signal\\$x$};
\node[box=green, below=7mm of x, minimum width=26mm] (p) {Fixed target\\$p^{*}$};

\node[box=blue, right=18mm of x, minimum width=52mm, minimum height=12mm] (G)
{Residual generator\\$\hat{x}=G(x,p^{*})$};
\node[box=blue, right=18mm of G, minimum width=22mm] (xhat) {Generated\\$\hat{x}$};

\draw[arrow] (x) -- (G);
\draw[arrow] (p.east) -- ++(0,2mm) -| (G.south);
\draw[arrow] (G) -- (xhat);

\node[box=gray, above=8mm of xhat, minimum width=40mm] (C)
{Classifier\\$p_{\mathrm{fault}}=f(\hat{x})$};
\draw[arrow] (xhat) -- (C);

\node[box=orange, below=10mm of G, minimum width=54mm, minimum height=12mm] (D)
{Discriminator\\$D(\cdot,p^{*})$};

\draw[arrow] (x.east) -- ++(10mm,0) |- (D.west);
\draw[arrow] (xhat.south) -- ++(0,-4mm) -| ([xshift=2mm]D.north);
\draw[arrow] (p.east) -- ++(0,-14mm) -| (D.south);

\node[loss=white, right=18mm of D, minimum width=70mm] (LG)
{\textbf{Generator loss}\\
$L_G=\alpha_{adv}L_{adv}+\lambda_{\mathrm{bce}}L_{prob}+\lambda_{res}L_{res}$\\
\quad $L_{adv}=-\mathbb{E}[D(\hat{x},p^{*})]$\\
\quad $L_{prob}=\mathrm{BCE}\!\left(p^{*},\, f(\hat{x})\right)$\\
\quad $L_{res}=\lVert\hat{x}-x\rVert_{1}$};

\node[loss=white, below=8mm of LG, minimum width=70mm] (LD)
{\textbf{Discriminator loss (WGAN-GP)}\\
$L_D=\mathbb{E}[D(\hat{x},p^{*})]-\mathbb{E}[D(x,p^{*})]+\lambda_{gp}L_{gp}$};

\draw[arrow] ([yshift=2mm]D.east) -- ([yshift=2mm]LG.west);
\draw[arrow] (C) -| (LG.north);

\draw[arrow] (D.east) -- ++(6mm,0) |- (LD.west);

\end{tikzpicture}
\end{adjustbox}

\caption{PR-GAN training schematic. A generator produces $\hat{x}$ from $x$ and $p^{*}$; the discriminator enforces realism and the classifier pushes $f(\hat{x})$ toward $p^{*}$.}
\label{fig:prgan_training}
\end{figure*}

The generator takes an existing sample $x$ as input and refines it using a residual formulation:
\begin{equation}
\hat{x} = G(x, p^{*}) = x + \Delta(x, p^{*}) .
\end{equation}
In this setup, the model learns a small adjustment to $x$ rather than generating a signal entirely from scratch. This design promotes small and controlled changes to the input sample, which helps preserve the underlying signal structure and maintain realism. So, we aim to apply minimal changes to the input sample when generating the output (Figure~\ref{fig:prgan_training}).
The pseudo-code of PR-GAN training algorithm is provided in Algorithm~\ref{alg:prgan}.

\begin{algorithm}[t]
\caption{PR-GAN training algorithm}
\label{alg:prgan}
\footnotesize
\begin{algorithmic}[1]
\Require Training set $X_{\text{train}}$, fixed classifier $f$, target probability $p^{*}$
\State Initialize generator $G$ and discriminator $D$; freeze $f$
\For{$epoch=1$ to $E$}
  \ForAll{$x \sim X_{\text{train}}$}
    \State Set conditioning target $p \leftarrow p^{*}$
    \For{$t=1$ to $n_c$}
      \State $\hat{x} \leftarrow G(x,p)$
      \State $L_D \leftarrow \mathbb{E}[D(\hat{x},p)]-\mathbb{E}[D(x,p)] + \lambda_{\text{GP}}\,GP(x,\hat{x},p)$
      \State Update $D$ by minimizing $L_D$
    \EndFor
    \State $\hat{x} \leftarrow G(x,p)$
    \State $L_{\text{adv}} \leftarrow -\mathbb{E}[D(\hat{x},p)]$
    \State $p_f \leftarrow f(\hat{x})$
    \State $L_G \leftarrow \alpha_{\text{adv}}L_{\text{adv}} + \lambda_{\mathrm{bce}}\,\mathrm{BCE}(p,p_f)
    + \lambda_{\text{res}}\,\mathbb{E}[\lVert \hat{x}-x\rVert_1]$
    \State Update $G$ by minimizing $L_G$
  \EndFor
\EndFor
\State \Return $G$
\end{algorithmic}
\end{algorithm}

\subsubsection{Discriminator}
The discriminator is optimized using the WGAN-GP objective:
\begin{equation}
L_D
=
\mathbb{E}\!\left[ D(\hat{x}, p^{*}) \right]
-
\mathbb{E}\!\left[ D(x, p^{*}) \right]
+
\lambda_{gp} \, L_{gp},
\end{equation}
where $\hat{x} = G(x, p^{*})$ and $L_{gp}$ is the gradient penalty computed on linear interpolations
\begin{equation}
\tilde{x} = x + \alpha \left( \hat{x} - x \right).
\end{equation}

\begin{equation}
L_{gp}
=
\mathbb{E}\!\left[
\left(
\left\lVert
\nabla_{\tilde{x}} D(\tilde{x}, p^{*})
\right\rVert_{2}
-
1
\right)^{2}
\right].
\end{equation}

\subsubsection{Generator}
The generator is trained with three terms: an adversarial term $L_{adv}$ for realism, a probability alignment term $L_{prob}$ that matches the frozen classifier’s fault probability to the target $p^{*}$, and an $L_{1}$ residual regularizer $L_{res}$ that limits unnecessary changes:

\begin{equation}
L_G
=
\alpha_{adv}\, L_{adv}
+
\lambda_{\mathrm{bce}}\, \mathbb{E}\!\left[\mathrm{BCE}\!\left(p^{*},\, f(\hat{x})\right)\right]
+
\lambda_{res}\, \mathbb{E}\!\left[ \lVert \hat{x} - x \rVert_{1} \right].
\end{equation}

where $f(\cdot)$ denotes the fault-class probability predicted by the classifier. In our implementation,
\begin{equation}
L_{adv}
=
-
\mathbb{E}\!\left[ D(\hat{x}, p^{*}) \right].
\end{equation}

\begin{equation}
\mathrm{BCE}\!\left(p^{*}, f(\hat{x})\right)
=
-\,p^{*}\log\!\big(f(\hat{x})\big)
-\left(1-p^{*}\right)\log\!\left(1-f(\hat{x})\right).
\end{equation}

\subsection{Counterfactuals (CF) method}
\label{sec:aecf}

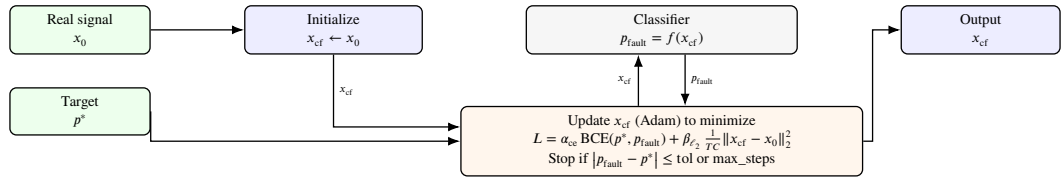
\begin{figure*}[t]
\centering
\begin{adjustbox}{width=0.80\textwidth}
\begin{tikzpicture}[
  font=\small,
  node distance=10mm and 18mm,
  box/.style={draw, rounded corners, thick, align=center, inner sep=5pt, fill=#1!7},
  arrow/.style={-{Latex[length=2.6mm]}, thick},
  looparrow/.style={-{Latex[length=2.6mm]}, thick, rounded corners}
]

\node[box=green, minimum width=30mm] (x0) {Real signal\\$x_0$};
\node[box=green, below=7mm of x0, minimum width=30mm] (pstar) {Target\\$p^{*}$};

\node[box=blue, right=20mm of x0, minimum width=38mm] (init)
{Initialize\\$x_{\mathrm{cf}}\leftarrow x_0$};

\node[box=gray, right=22mm of init, minimum width=58mm] (C)
{Classifier\\$p_{\mathrm{fault}}=f(x_{\mathrm{cf}})$};

\node[box=blue, right=22mm of C, minimum width=34mm] (out)
{Output\\$x_{\mathrm{cf}}$};

\node[box=orange, below=11mm of C, minimum width=86mm] (U)
{Update $x_{\mathrm{cf}}$ (Adam) to minimize\\
$L=\alpha_{\mathrm{ce}}\,\mathrm{BCE}(p^{*},p_{\mathrm{fault}})+\beta_{\ell_2}\,\frac{1}{TC}\lVert x_{\mathrm{cf}}-x_0\rVert_2^{2}$\\
Stop if $\left|p_{\mathrm{fault}}-p^{*}\right|\le\mathrm{tol}$ or max\_steps};

\draw[arrow] (x0) -- (init);
\draw[arrow] (init.south) |- node[right, pos=0.25]{\scriptsize $x_{\mathrm{cf}}$} ([yshift=5mm]U);
\draw[arrow] (U.east) -- ++(2mm,0) |- (out.west);

\draw[arrow] ([xshift=5mm]C.south) -- node[right]{\scriptsize $p_{\mathrm{fault}}$} ([xshift=5mm]U.north);
\draw[arrow] (pstar.east) |- (U.west);

\draw[arrow] ([xshift=-5mm]U.north) -- node[left]{\scriptsize $x_{\mathrm{cf}}$} ([xshift=-5mm]C.south);

\end{tikzpicture}
\end{adjustbox}
\caption{Counterfactual (CF) method. Starting from a real signal \(x_0\), we iteratively update \(x_{\mathrm{cf}}\) so that a classifier outputs \(p_{\mathrm{fault}}=f(x_{\mathrm{cf}})\) close to the target \(p^{*}\), while keeping \(x_{\mathrm{cf}}\) close to \(x_0\).}
\label{fig:cf_generation_simple}
\end{figure*}

Our second approach is a counterfactual procedure based on direct optimization at inference time. The form of the objective follows Wachter et al.~\cite{wachter2018gdpr}: we minimize a weighted sum of a loss that pushes the classifier's prediction toward a target and an $L_2$ proximity term that keeps the modified input close to the source. Two adaptations are specific to our setting. First, the target is a continuous probability value $p^{*}\in\{0.25,0.50,0.75\}$ rather than a class label, so the prediction-matching term is a binary cross-entropy with a soft target rather than a class-flip loss. Second, the input is a multi-channel bearing vibration window, and the proximity term and amplitude clipping are applied accordingly. Starting from an input sample $x_0$ and a desired probability $p^{*}$, the method updates the signal with small steps so that the classifier's predicted fault probability moves toward the target while the signal stays close to the original sample. The optimization stops when the target is reached within a tolerance or when a maximum number of steps is met (Figure~\ref{fig:cf_generation_simple}). The maximum number of steps is set to 300. We set the stopping tolerance to \(\epsilon=0.05\). The pseudo-code is provided in Algorithm~\ref{alg:cf}.

\begin{algorithm}[t]
\caption{Counterfactual Algorithm}
\label{alg:cf}
\footnotesize
\begin{algorithmic}[1]
\Require Input signal $x_0$, fixed classifier $f$, target fault probability $p^{*}$
\State Initialize counterfactual $\hat{x} \gets x_0$
\For{$s=1$ to $S$}
  \State $p_f \gets f(\hat{x})$
    \State $L \gets \alpha_{\mathrm{ce}}\,\mathrm{BCE}(p^{*}, p_f) + \beta_{\ell_2}\,\frac{1}{TC}\lVert \hat{x}-x_0\rVert_2^2$
  \State Update $\hat{x}$ by an Adam step to minimize $L$
  \State Clip $\hat{x}$ to a valid amplitude range
  \State $p_f \gets f(\hat{x})$
  \If{$|p_f - p^{*}| \le \epsilon$}
    \State \textbf{break}
  \EndIf
\EndFor
\State \Return $\hat{x}$
\end{algorithmic}
\end{algorithm}

\paragraph{Optimization objective.}
In this design, we optimize the counterfactual sample $x_{\mathrm{cf}}$ by minimizing the following objective:
\begin{equation}
L_{CF}(x_{\mathrm{cf}}; x_0, p^{*})
=
\alpha_{\mathrm{ce}} \, L_{\mathrm{ce}}(x_{\mathrm{cf}}, p^{*})
+
\beta_{\ell_2} \, L_{\ell_2}(x_{\mathrm{cf}}, x_0).
\end{equation}

The term $L_{\mathrm{ce}}$ is a binary cross-entropy between the target probability $p^{*}$ and the classifier’s predicted fault probability $f(x_{\mathrm{cf}})$:
\begin{equation}
L_{\mathrm{ce}}(x_{\mathrm{cf}}, p^{*})
=
\mathrm{BCE}\!\left( p^{*}, f(x_{\mathrm{cf}}) \right).
\end{equation}

To keep the counterfactual signal close to the original real sample $x_0$, we use an $L_2$ distance term that penalizes deviations:
\begin{equation}
L_{\ell_2}(x_{\mathrm{cf}}, x_0)
=
\frac{1}{TC}
\left\lVert x_{\mathrm{cf}} - x_0 \right\rVert_2^2.
\end{equation}

This objective encourages the optimizer to reach the target probability with minimal and realistic changes to the input signal.

\subsection{Evaluation metrics}
In this section, we evaluate the generated samples from two perspectives:
\begin{itemize}
    \item \textbf{Target probability alignment:} we measure how closely fault probabilities of generated samples match the target probability.
    \item \textbf{Similarity to the original source:} we evaluate how similar each generated signal remains to its corresponding original sample by comparing them in the time domain and the frequency domain.
\end{itemize}
We use the source-similarity metrics ($\Delta L_1$, $\Delta\mathrm{TV}$, $d_{\mathrm{PSD}}$) as proxies for realism, on the assumption that generated signals close to a real source sample are more plausible than those that depart far from it. We note the limitation of this design: a method that simply returns the source signal unchanged scores zero on all three similarity metrics while failing the probability-alignment objective. The metrics therefore measure how aggressively a method edits the input rather than the realism of the result in any absolute sense, and they should be read jointly with the probability-error and success-rate metrics. A more direct realism evaluation, such as envelope-spectrum analysis at characteristic bearing fault frequencies or a learned realism classifier, is left to future work.

Firstly, we filtered the generated samples using a target-probability error threshold of 0.25; samples with \(|p_{\mathrm{fault}}(x_i)-p^{*}| > 0.25\) were removed and counted as failed or outliers. This filtering step is applied because samples that are far from the target probability do not satisfy the goal of probability-targeted generation and could bias the reported evaluation metrics. We computed the mean and standard deviation of the remaining fault probabilities and evaluated the retained samples using the metrics below. We note that all metrics computed after filtering, including the success rate $\mathrm{SR}_{\tau}$, are conditioned on retained samples and should be read alongside $N_{failed}$ rather than in isolation. A method that fails on a substantial fraction of inputs but achieves a high $\mathrm{SR}_{\tau}$ on the retained ones is therefore not directly comparable to one that produces fewer failures.

\paragraph{Standard deviation (STD)}
We computed the standard deviation (STD) of the generated signals and their source signals to evaluate signal diversity:

\begin{equation}
\mu_{t,c}
=
\frac{1}{N}\sum_{i=1}^{N} X_{i,t,c}.
\end{equation}

\begin{equation}
\mathrm{STD}
=
\sqrt{
\frac{1}{NTC}
\sum_{i=1}^{N}
\sum_{t=1}^{T}
\sum_{c=1}^{C}
\left(X_{i,t,c}-\mu_{t,c}\right)^2
}.
\end{equation}

\paragraph{Target-probability error (MAE)}
For each generated signal $x_i$, we measure the deviation between the classifier’s predicted fault probability and the target probability $p^{*}$. This metric shows how well a generation method can produce samples with the desired fault probability:
\begin{equation}
MAE_p = \frac{1}{M} \sum_{i=1}^{M} |p_{\mathrm{fault}}(x_i) - p^{*}| ,
\end{equation}
where $M$ is the number of retained generated samples.

\paragraph{Success rate within tolerance ($\mathrm{SR}_{\tau}$).}
We report the proportion of generated samples whose predicted fault probability is within a tolerance $\tau$ of the target. For each generated sample $x_i$ with predicted probability $p_{\mathrm{fault}}(x_i)$, we define

\begin{equation}
\mathrm{SR}_{\tau}
=
\frac{1}{M}
\sum_{i=1}^{M}
\mathbf{1}\!\left(
\left|p_{\mathrm{fault}}(x_i)-p^{*}\right|
\le
\tau
\right),
\end{equation}
where $\mathbf{1}(\cdot)$ is the indicator function, $\tau$ corresponds to the tolerance parameter which is $0.05$, and $M$ is the number of retained generated samples.

\paragraph{$\Delta L_{1}$ (mean absolute difference in the time domain).}
To measure how much a generated signal deviates from its corresponding source signal in the time domain, we compute the mean absolute difference between the generated sample $x_i$ and its source sample $x^{\mathrm{src}}_i$:
\begin{equation}
\Delta L_{1}(\hat{x}_i, x_i^{\mathrm{src}})
=
\frac{1}{TC}
\sum_{t=1}^{T}
\sum_{c=1}^{C}
\left|
\hat{x}_i(t,c) - x_i^{\mathrm{src}}(t,c)
\right|.
\end{equation}
We then report the average over all samples:
\begin{equation}
\overline{\Delta L_{1}}
=
\frac{1}{M}
\sum_{i=1}^{M}
\Delta L_{1}(\hat{x}_i, x_i^{\mathrm{src}}),
\end{equation}
where $T$ is the signal length, $C$ is the number of channels, and $M$ is the number of samples.

\paragraph{Total variation difference ($\Delta \mathrm{TV}$).}
To quantify time-domain realism and detect over-smoothing or local artifacts, we compute the average absolute first difference and report the absolute difference between the generated signal and its paired source signal:
\begin{equation}
\mathrm{TV}(x)
=
\frac{1}{(T-1)C}
\sum_{t=1}^{T-1}
\sum_{c=1}^{C}
\left| x_{t+1,c} - x_{t,c} \right|,
\end{equation}
\begin{equation}
\Delta \mathrm{TV}_i
=
\left|
\mathrm{TV}(\hat{x}_i)
-
\mathrm{TV}(x_i^{\mathrm{src}})
\right|.
\end{equation}
\begin{equation}
\overline{\Delta \mathrm{TV}}
=
\frac{1}{M}
\sum_{i=1}^{M}
\Delta \mathrm{TV}_i .
\end{equation}
where $M$ is the number of evaluated samples (after probability-threshold filtering).

\paragraph{Spectral difference (PSD).}
Because bearing faults are strongly reflected in the frequency domain, we compare the power spectral density (PSD) of each generated sample to that of its source sample. For each sample pair $(\hat{x}_i, x_i^{\mathrm{src}})$, we estimate the PSD using Welch's method and compute the log-PSD mean squared error (MSE) across frequency bins. For each channel $c$ we define
\begin{equation}
\begin{aligned}
d_{\mathrm{PSD}}^{(c)}(\hat{x}_i, x_i^{\mathrm{src}})
&=
\frac{1}{|F|}
\sum_{f \in F}
\Bigl(
\log\!\bigl(\mathrm{PSD}(\hat{x}_i; f,c)\bigr)
\\
&\qquad\qquad
-
\log\!\bigl(\mathrm{PSD}(x_i^{\mathrm{src}}; f,c)\bigr)
\Bigr)^2 .
\end{aligned}
\end{equation}
and average across channels:
\begin{equation}
d_{\mathrm{PSD}}(\hat{x}_i, x_i^{\mathrm{src}})
=
\frac{1}{C}
\sum_{c=1}^{C}
d_{\mathrm{PSD}}^{(c)}(\hat{x}_i, x_i^{\mathrm{src}}).
\end{equation}
We then report the mean spectral difference over all samples:
\begin{equation}
\overline{d_{\mathrm{PSD}}}
=
\frac{1}{M}
\sum_{i=1}^{M}
d_{\mathrm{PSD}}(\hat{x}_i, x_i^{\mathrm{src}}),
\end{equation}
where $F$ is the set of frequency bins, $C$ is the number of channels.

\section{Results and Discussion}
\label{sec:results_discussion}
This section reports the results of our experiments. We evaluate the baseline classifier, PR-GAN, and the counterfactual method for generating new samples with target fault probabilities of $0.25$, $0.50$, and $0.75$. We also present the data distribution for each studied class in this section.

\subsection{Baseline classifier}
The baseline classifier achieves reliable performance on the test split for both the CWRU and Paderborn datasets, as reflected by its accuracy and F1-score (Table~\ref{tab:baseline_classifier_performance}). However, our aim is to not examine and analyze its results deeply, and we show some details and results of our model just for presentation purposes.

\begin{table}[t]
\centering
\caption{Baseline classifier performance on the binary normal--fault classification task.}
\label{tab:baseline_classifier_performance}
\begin{tabular}{lccc}
\hline
\textbf{Dataset} & \textbf{Test Accuracy} & \textbf{Normal F1} & \textbf{Fault F1} \tabularnewline
\hline
CWRU & 0.99 & 0.99 & 0.99 \tabularnewline
Paderborn & 0.86 & 0.88 & 0.85 \tabularnewline
\hline
\end{tabular}
\end{table}

\subsection{Probability Distribution of Data}
\begin{figure*}[t]
\centering

\begin{subfigure}{0.48\linewidth}
\centering
\includegraphics[width=\linewidth]{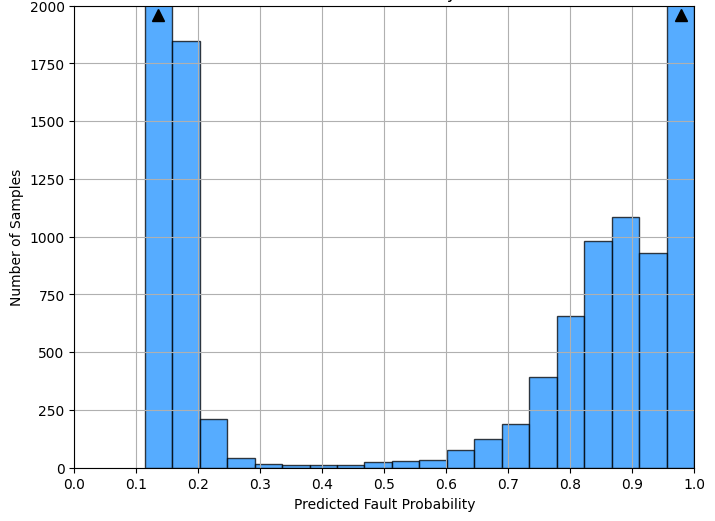}
\caption{Distribution of CWRU dataset.}
\label{fig:cwru_distribution}
\end{subfigure}
\hfill
\begin{subfigure}{0.48\linewidth}
\centering
\includegraphics[width=\linewidth]{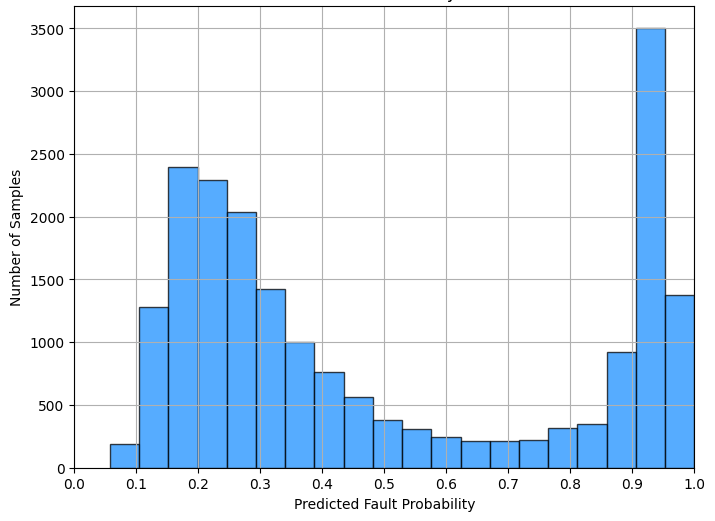}
\caption{Distribution of Paderborn dataset.}
\label{fig:paderborn_distribution}
\end{subfigure}

\caption{Distribution of predicted fault probabilities for the CWRU and Paderborn datasets.}
\label{fig:datasets_distribution}
\end{figure*}

\begin{figure*}[t]
\centering

\begin{subfigure}{0.48\linewidth}
\centering
\includegraphics[width=\linewidth]{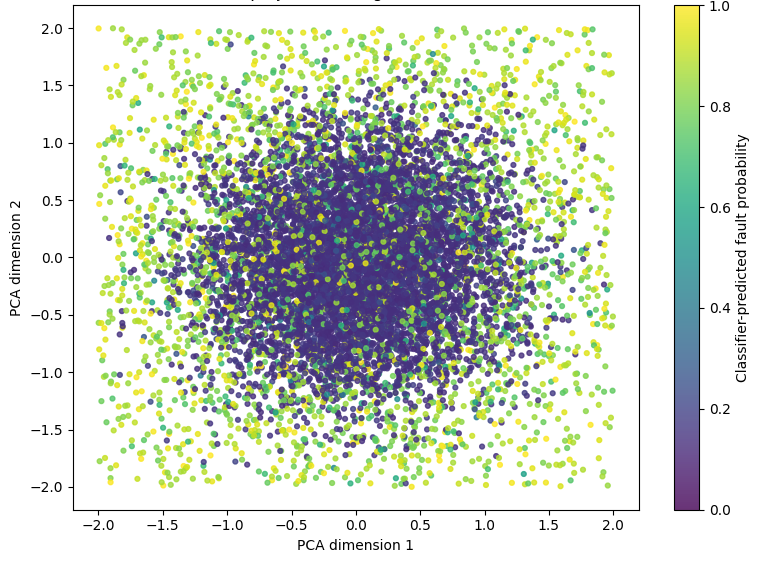}
\caption{CWRU dataset.}
\label{fig:cwru_pca}
\end{subfigure}
\hfill
\begin{subfigure}{0.48\linewidth}
\centering
\includegraphics[width=\linewidth]{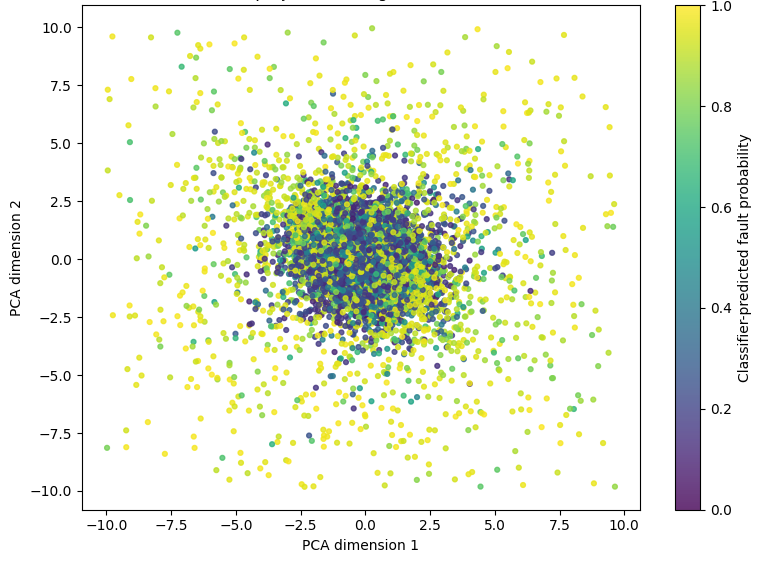}
\caption{Paderborn dataset.}
\label{fig:paderborn_pca}
\end{subfigure}

\caption{PCA projection of time-domain features, colored by the predicted fault probability.}
\label{fig:datasets_pca}
\end{figure*}

We use the baseline classifier to estimate the fault-probability distributions for the CWRU and Paderborn datasets. In our experiments, we assume that the classifier provides meaningful probability estimates. The results are reported in Table~\ref{tab:prob_distribution_datasets} and in Figure~\ref{fig:datasets_distribution}. Many samples receive probabilities close to 0 or 1, which indicates fewer mid-range (gray-zone) cases in some intervals, roughly in the $0.2$--$0.8$ fault-probability interval. Some probability intervals contain a relatively large number of samples, whereas some other intervals contain fewer samples in this range. 
Also, in Fig.~\ref{fig:datasets_pca}, the PCA projection of time-domain features complements the probability histograms by showing how the classifier-predicted fault probabilities are distributed in the signal feature space.

\begin{table}[t]
\centering
\caption{Distribution of predicted fault probabilities for the CWRU and Paderborn datasets.}
\label{tab:prob_distribution_datasets}
\resizebox{\linewidth}{!}{%
\begin{tabular}{lrrrr}
\hline
 & \multicolumn{2}{c}{CWRU} & \multicolumn{2}{c}{Paderborn} \\
\cline{2-3} \cline{4-5}
Probability bin ($p_{\mathrm{fault}}$)
& count & percent
& count & percent \\
\hline
$[0.0, 0.1)$ &    0 &  0.00\% &  141 &  0.70\% \\
$[0.1, 0.2)$ & 6460 & 47.57\% & 3793 & 18.96\% \\
$[0.2, 0.3)$ &  301 &  2.22\% & 4537 & 22.68\% \\
$[0.3, 0.4)$ &   45 &  0.33\% & 2396 & 11.98\% \\
$[0.4, 0.5)$ &   38 &  0.28\% & 1255 &  6.28\% \\
$[0.5, 0.6)$ &   70 &  0.52\% &  657 &  3.28\% \\
$[0.6, 0.7)$ &  218 &  1.61\% &  451 &  2.25\% \\
$[0.7, 0.8)$ &  742 &  5.46\% &  527 &  2.63\% \\
$[0.8, 0.9)$ & 2321 & 17.09\% & 1115 &  5.58\% \\
$[0.9, 1.0]$ & 3385 & 24.93\% & 5128 & 25.64\% \\
\hline
\textbf{Total} & \textbf{13580} & \textbf{100.00\%}
               & \textbf{20000} & \textbf{100.00\%} \\
\hline
\end{tabular}%
}
\end{table}

\subsection{Data generation for CWRU dataset}
We report generated samples and their evaluations and comparisons for the CWRU dataset with three different fault probabilities in this part. We use PR-GAN and counterfactual methods for data generation here.

The results are reported in Table~\ref{tab:metrics_targets_cwru}. The results are reported in Table~\ref{tab:metrics_targets_cwru}. Overall, the counterfactual (CF) method provides more accurate probability steering, whereas PR-GAN exhibits higher variability with more failed samples. Across all target probabilities, CF achieves tight control of the target probability, with a mean absolute target error of approximately $0.005$--$0.008$. In contrast, PR-GAN’s target error is larger and less stable, ranging from $0.048$ to $0.059$, and it has more failed generated samples. CF also better preserves some signal characteristics: it yields lower log-PSD differences and lower mean absolute time-domain differences across all settings, although its total variation differences are higher than those of PR-GAN. Also, PR-GAN has a lower reported runtime on this dataset. PR-GAN usually stays closer to the STD of the original data. Also, as the target probability increases, the internal variation slightly increases for both methods, particularly for PR-GAN.
Qualitative examples of real and generated samples for the inner race fault class are shown in Fig.~\ref{fig:fig-cwru}.

\begin{table*}[t]
\centering
\caption{Evaluation metrics for generated samples at different target fault probabilities $p^{*}$ on the CWRU dataset.}
\label{tab:metrics_targets_cwru}
\vspace{0.5em}

\begin{threeparttable}

\resizebox{\textwidth}{!}{%
\begin{tabular}{r l r r r r r r r r r r r}
\hline
Target ($p^{*}$) & Method & N$*{\mathrm{failed}}$ & Mean$*p$ & STD$*p$ & MAE$*p$ & SR$*{\tau}$ & $\Delta L*{1}$ & $\Delta TV$ & $d*{\text{PSD}}$ & Time (minute)\tnote{1} & STD$*{\mathrm{generated}}$ & STD$_{\mathrm{source}}$ \tabularnewline
\hline
\textbf{0.25} & PR-GAN & 77 & 0.227 & 0.070 & 0.059 & 0.501 & 0.130 & 0.017 & 0.537 & 43.76  & 0.170 & 0.177 \tabularnewline
& CF     & 0  & 0.253 & 0.012 & 0.008 & 1.000 & 0.067 & 0.046 & 0.370 & 213.58 & 0.189 & 0.233 \tabularnewline
\cline{1-13}
\textbf{0.50} & PR-GAN & 74 & 0.498 & 0.074 & 0.052 & 0.624 & 0.123 & 0.019 & 0.518 & 46.76  & 0.179 & 0.180 \tabularnewline
& CF     & 0  & 0.501 & 0.006 & 0.005 & 1.000 & 0.060 & 0.040 & 0.257 & 115.30 & 0.197 & 0.233 \tabularnewline
\cline{1-13}
\textbf{0.75} & PR-GAN & 3  & 0.769 & 0.067 & 0.048 & 0.680 & 0.128 & 0.024 & 0.529 & 44.05  & 0.234 & 0.234 \tabularnewline
& CF     & 0  & 0.753 & 0.006 & 0.006 & 1.000 & 0.056 & 0.043 & 0.242 & 127.51 & 0.201 & 0.233 \tabularnewline
\hline
\end{tabular}%
}

\begin{tablenotes}
\footnotesize
\item[1] Time denotes the total training time for PR-GAN and the total generation time for the CF method.
\end{tablenotes}

\end{threeparttable}
\end{table*}

\subsection{Data generation for Paderborn dataset}
In this part, we report the samples generated by PR-GAN and the counterfactual (CF) method, along with their evaluation results, for the Paderborn dataset at three target probabilities.

Table~\ref{tab:metrics_targets_paderborn} reports the generation results on the Paderborn dataset at target probabilities $p^{*} \in \{0.25, 0.50, 0.75\}$. The counterfactual (CF) method is more reliable than PR-GAN in steering samples toward the target probability and in preserving signal characteristics in most settings. PR-GAN has more failed generated samples. In contrast, CF produces samples closer to the target and achieves lower target-probability errors. CF also more consistently reduces log-PSD differences and mean absolute time-domain differences, although it yields higher total-variation differences relative to the corresponding source signals. Overall, similar to what we observed on the CWRU dataset, these results suggest that CF provides a more stable approach for generating intermediate-probability samples. Also, the reported total runtime is comparable for the two methods, with CF being slightly higher overall on this dataset. Both PR-GAN and CF usually show lower internal variation than the original data, and this is clearer and more consistent for CF. In contrast, PR-GAN is generally closer to the diversity level of the original data.
Qualitative examples of real and generated signals for the real inner ring fault class are shown in Fig.~\ref{fig:fig-paderborn}.

\begin{table*}[t]
\centering
\caption{Evaluation metrics for generated samples at different target fault probabilities $p^{*}$ on the Paderborn dataset.}
\label{tab:metrics_targets_paderborn}
\vspace{0.5em}

\begin{threeparttable}

\resizebox{\textwidth}{!}{%
\begin{tabular}{r l r r r r r r r r r r r}
\hline
Target ($p^{*}$) & Method & N$_{\mathrm{failed}}$ & Mean$_p$ & STD$_p$ & MAE$_p$ & SR$_{\tau}$ & $\Delta L_{1}$ & $\Delta TV$ & $d_{\text{PSD}}$ & Time (minute)\tnote{1} & STD$_{\mathrm{generated}}$ & STD$_{\mathrm{source}}$ \tabularnewline
\hline
\textbf{0.25} & PR-GAN & 20 & 0.248 & 0.062 & 0.046 & 0.650 & 0.109 & 0.006 & 0.446 & 63.53 & 0.282 & 0.293 \tabularnewline
              & CF     & 0  & 0.251 & 0.006 & 0.005 & 1.000 & 0.057 & 0.015 & 0.243 & 82.81 & 0.257 & 0.323 \tabularnewline
\cline{1-13}
\textbf{0.50} & PR-GAN & 15 & 0.523 & 0.070 & 0.058 & 0.509 & 0.117 & 0.005 & 0.447 & 77.34 & 0.292 & 0.298 \tabularnewline
              & CF     & 0  & 0.500 & 0.006 & 0.005 & 1.000 & 0.051 & 0.016 & 0.167 & 80.68 & 0.267 & 0.323 \tabularnewline
\cline{1-13}
\textbf{0.75} & PR-GAN & 0  & 0.763 & 0.063 & 0.047 & 0.640 & 0.114 & 0.006 & 0.411 & 83.99 & 0.319 & 0.323 \tabularnewline
              & CF     & 0  & 0.751 & 0.006 & 0.005 & 1.000 & 0.051 & 0.016 & 0.135 & 76.63 & 0.271 & 0.323 \tabularnewline
\hline
\end{tabular}%
}

\begin{tablenotes}
\footnotesize
\item[1] Time denotes the total training time for PR-GAN and the total generation time for the CF method.
\end{tablenotes}

\end{threeparttable}
\end{table*}

\subsection{Discussion}

\subsubsection{Training-based vs. training-free approaches}
Training-based methods (e.g., PR-GAN) must be trained before generating data. This training stage takes time, and the target properties of the generated samples are learned during training. Once trained, however, generation is typically fast. A limitation is that the generator does not explicitly verify, at inference time, that each generated sample satisfies the desired conditions, which increases the risk of missing the target in some cases. Also, after any change in the settings, the model should be trained again. The performance of these models depends on the training settings and the inputs to the models. The training time primarily depends on the size of the dataset and the per-sample generation time is negligible.

In contrast, training-free methods do not require a separate training stage. Instead, they operate on each input sample by directly optimizing it and checking the target condition during the process. In the specific configuration we use, with 300 Adam steps and an early-stopping tolerance, the per-sample optimization is given a generous step budget against a direct objective; this is what makes it more reliable in reaching the target in our experiments, and the same property would not necessarily hold for shorter step budgets or different objectives. The price is that generation is performed separately for each sample and is therefore slower per sample. The counterfactual method applies small changes to the input signal to produce an output that remains close to the original sample while meeting the target probability. Any change in settings can be applied instantly to subsequent generated samples, and the generated samples can be independent of each other. The results depend on the settings, the starting points (inputs), and the guidance classifier. Generation time depends on the input sample.

Choosing between training-based and training-free approaches depends on practical and data-dependent factors. A key consideration is where computation is preferred, during an offline training phase or during per-sample generation. The input type also matters: when the output should remain closely tied to a specific real input sample and its structure, or when there is a sample-specific relation between the input and the desired output, which can vary from one input to another. A training-free method can be a better fit because it optimizes each input individually.

\paragraph{Computational cost.} The cost of the two methods is best reported as three separate quantities rather than as a single ``time per run.'' First, PR-GAN training time, paid once per setting: roughly 44--47 minutes on CWRU and 64--84 minutes on Paderborn. Second, PR-GAN inference time per sample once trained: a single forward pass through the generator, on the order of a fraction of a second, so the per-sample cost at inference is effectively negligible. Third, CF generation time per sample: each input is optimized independently for up to 300 Adam steps, so the 500-sample batches in our tables take 76--214 minutes overall, i.e. approximately 9--26 seconds per sample. PR-GAN therefore amortizes a large up-front training cost across an arbitrary number of subsequent generations, while CF pays its cost on every new sample. Which is preferable depends on how many samples will be generated for a given setting and on whether settings are likely to change between runs.

\paragraph{Confounders in the comparison.} We note that the PR-GAN/CF comparison is not a controlled ablation. The two methods differ along several axes simultaneously: training-time versus inference-time optimization, presence or absence of an adversarial realism term, residual editing versus direct signal optimization, and different effective optimization budgets per sample. Our results characterize the two methods as we have implemented them and should not be read as isolating the effect of any single one of these axes. In particular, the per-sample CF advantage on probability alignment partly reflects that CF is given 300 Adam steps with a direct probability objective at inference, whereas PR-GAN must amortize over inputs.

\subsubsection{PR-GAN vs. Counterfactual data generation}
In our experiments, the counterfactual method was generally more reliable and stable in reaching the target probability and in preserving signal characteristics. It achieved lower target-probability error and was generally better at preserving signal characteristics in terms of mean absolute time-domain difference and PSD distance, although not in terms of TV. In contrast, PR-GAN showed larger variation and had more failed generated samples in some cases.

\subsubsection{What does $p^{*}=0.5$ mean for near-degenerate or bimodal classifiers?}
The probability distributions in Table~\ref{tab:prob_distribution_datasets} is uneven in the binary normal--fault setting, and this affects how a generated sample at $p^{*}=0.5$ should be interpreted. Asking the classifier to assign $p_{\mathrm{fault}}(\hat{x})=0.5$ to a generated sample is asking for an output that the classifier may rarely produce on real data. The methods can still achieve this output, especially CF, which pushes per-sample probabilities to the target with high reliability. We caution that such samples may lie closer to adversarial perturbations of the source than to physically meaningful borderline conditions: they sit in regions of input space that the classifier has rarely encountered, and the classifier's $0.5$ output there is best read as ``locally uncertain on this perturbation'' rather than as a calibrated 50\% fault likelihood. We note this as a limitation of the present study and a direction for follow-up work that explicitly evaluates the on-manifold plausibility of generated borderline samples.

\section{Conclusion}
\label{sec:conclusion}
In this paper, we introduced two approaches for probability-targeted data generation of bearing vibration signals on the CWRU and Paderborn datasets with the goal of matching a user-specified fault probability. The first is PR-GAN, a training-based method that combines WGAN-GP with a probability-alignment term and a residual editing formulation. The second is a Wachter-style counterfactual procedure, training-free and per-sample, applied to vibration windows with a real-valued probability target.

PR-GAN requires a training stage but, once trained, generates each sample in a single forward pass and is therefore fast at inference. CF updates each input iteratively at generation time, which makes it slower per sample. In our experiments, CF achieved tighter probability alignment (mean absolute error of $0.005$--$0.008$, success rate close to $1.00$ on retained samples) and lower mean absolute time-domain difference and log-PSD distance, while PR-GAN had larger probability error ($0.046$--$0.059$ on retained samples), more failed generations, and more variability across settings.

Our results show that intermediate-probability vibration samples can be synthesized while keeping the generated signal close to a real source sample. We do not claim that all such samples are physically meaningful: as discussed in Section~\ref{sec:results_discussion}, datasets or settings whose predicted-probability distribution is sharply bimodal or near-degenerate are likely to produce target-probability samples that lie closer to adversarial perturbations than to plausible borderline conditions. With that caveat, probability-targeted samples provide a practical way to populate the boundary region of a fixed classifier with controlled $p^{*}$ values that are otherwise scarce in standard bearing datasets.

A central limitation of the present study is that we did not analyze the calibration of the guidance ensemble. The interpretation of any specific generated sample as ``a signal at fault probability $p^{*}$'' is only as meaningful as the classifier's probability output, and we have not measured expected calibration error or reliability diagrams here. We made this scoping choice to isolate the behavior of the two generation procedures from the behavior of the classifier, but it leaves open the question of whether the synthesized samples are useful in any specific downstream task such as calibration evaluation, out-of-distribution detection, or augmentation for boundary robustness. Future work should (i) quantify the calibration of the guidance ensemble, (ii) repeat the experiments with alternative guidance models including calibrated ones, (iii) add a direct realism evaluation such as envelope-spectrum analysis at characteristic bearing fault frequencies, and (iv) demonstrate at least one downstream use of the generated borderline samples.

\appendices

\section{Additional visual results}
\label{app:qualitative}

\begin{figure*}[t]
\centering

\begin{subfigure}[t]{\textwidth}
\centering
\begin{subfigure}[t]{0.49\textwidth}
\centering
\includegraphics[width=\linewidth]{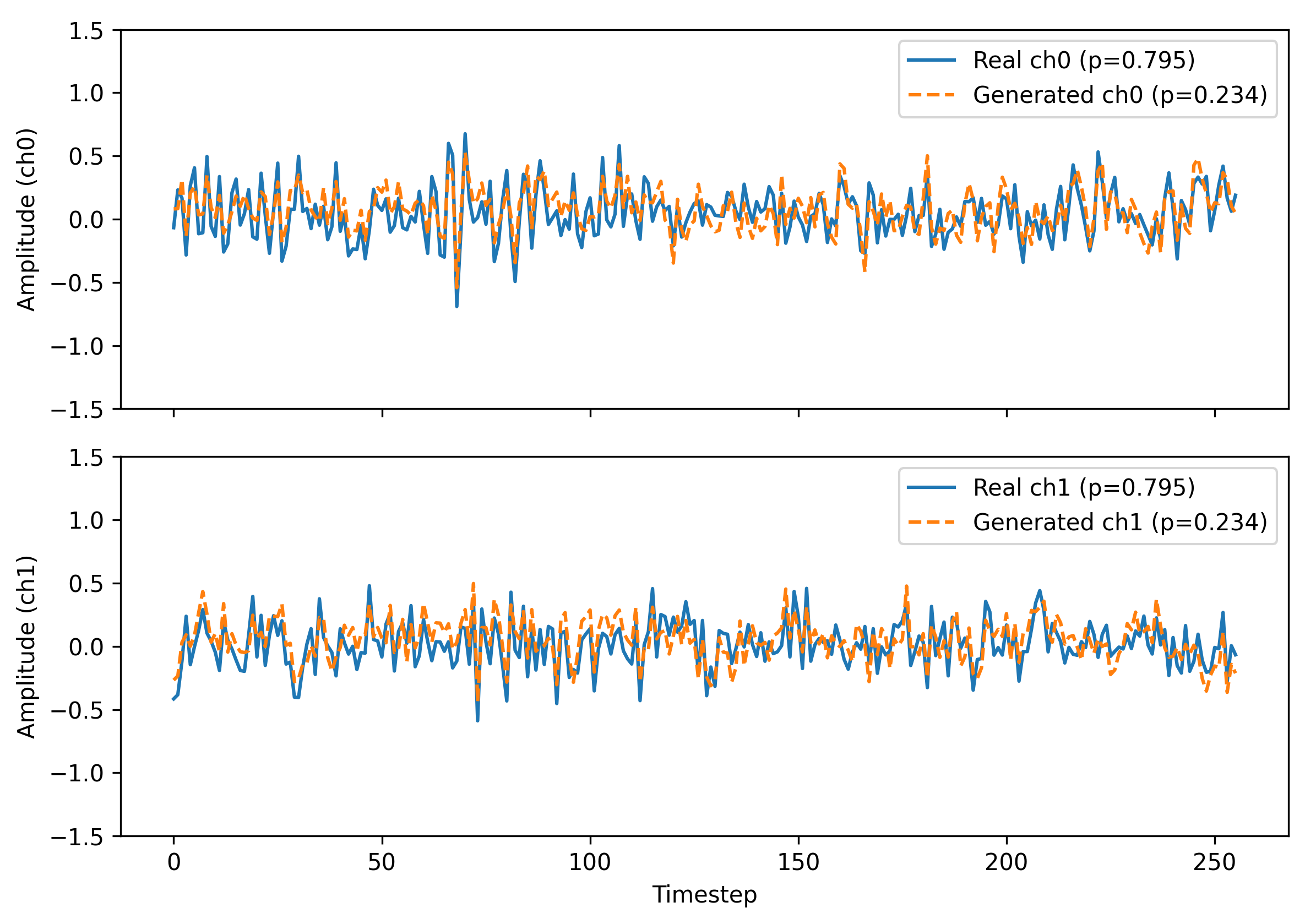}
\caption{PR-GAN generated sample vs. original sample}
\label{fig:fig-cwru-025-gan}
\end{subfigure}
\hfill
\begin{subfigure}[t]{0.49\textwidth}
\centering
\includegraphics[width=\linewidth]{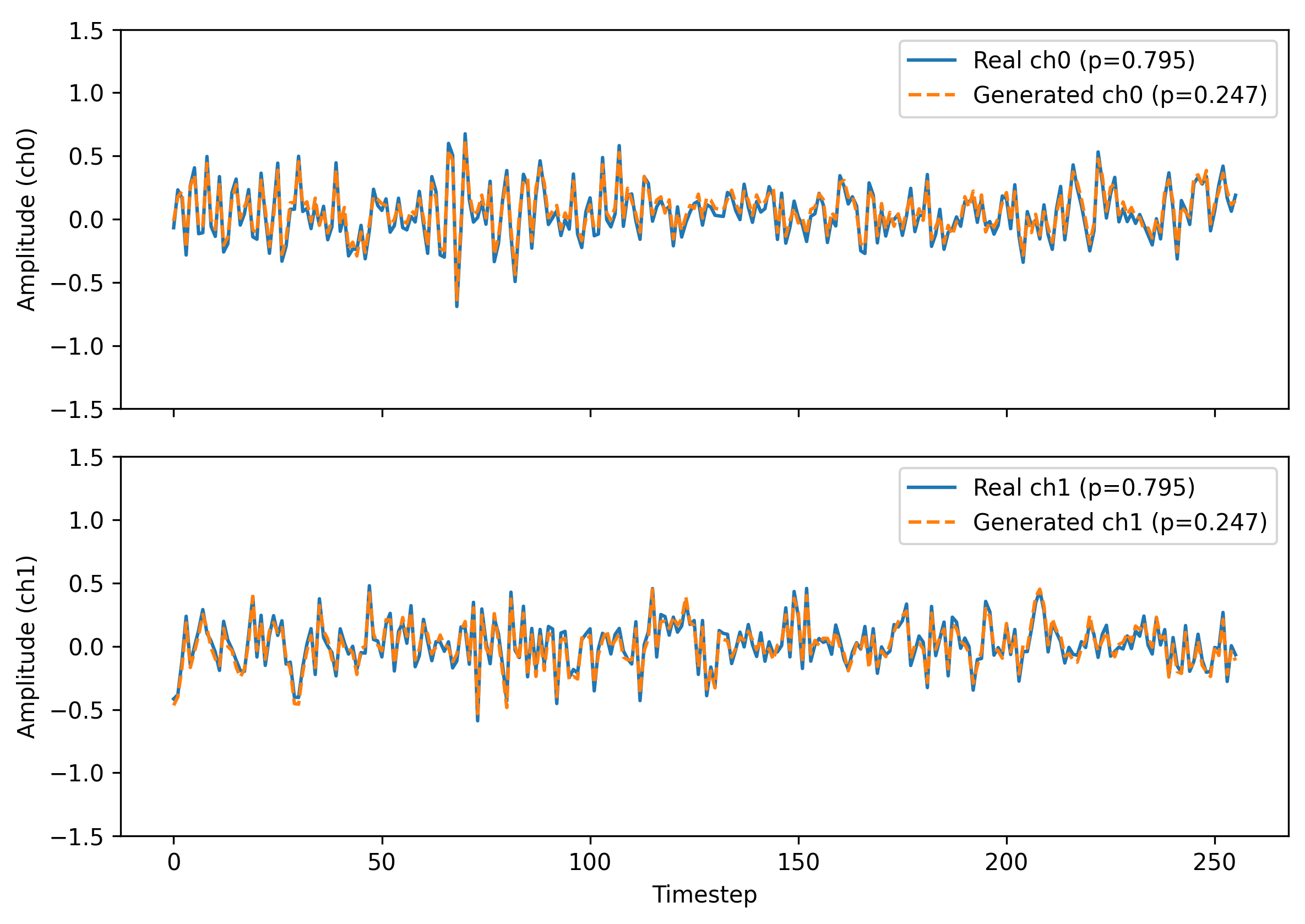}
\caption{Counterfactual generated sample vs. original sample}
\label{fig:fig-cwru-025-cf}
\end{subfigure}
\caption{Target probability 0.25}
\label{fig:fig-cwru-025}
\end{subfigure}

\vspace{0.6em}

\begin{subfigure}[t]{\textwidth}
\centering
\begin{subfigure}[t]{0.49\textwidth}
\centering
\includegraphics[width=\linewidth]{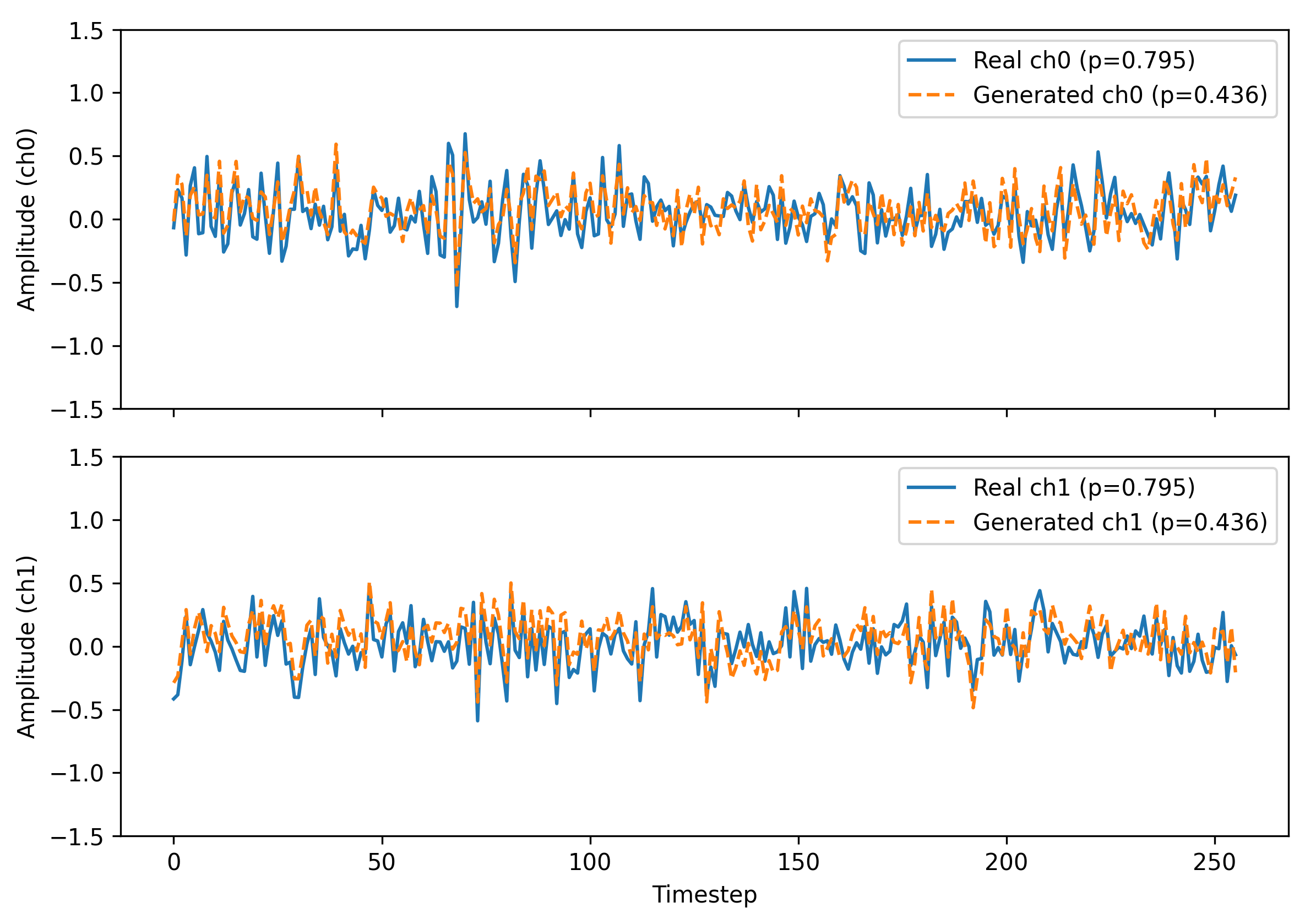}
\caption{PR-GAN generated sample vs. original sample}
\label{fig:fig-cwru-05-gan}
\end{subfigure}
\hfill
\begin{subfigure}[t]{0.49\textwidth}
\centering
\includegraphics[width=\linewidth]{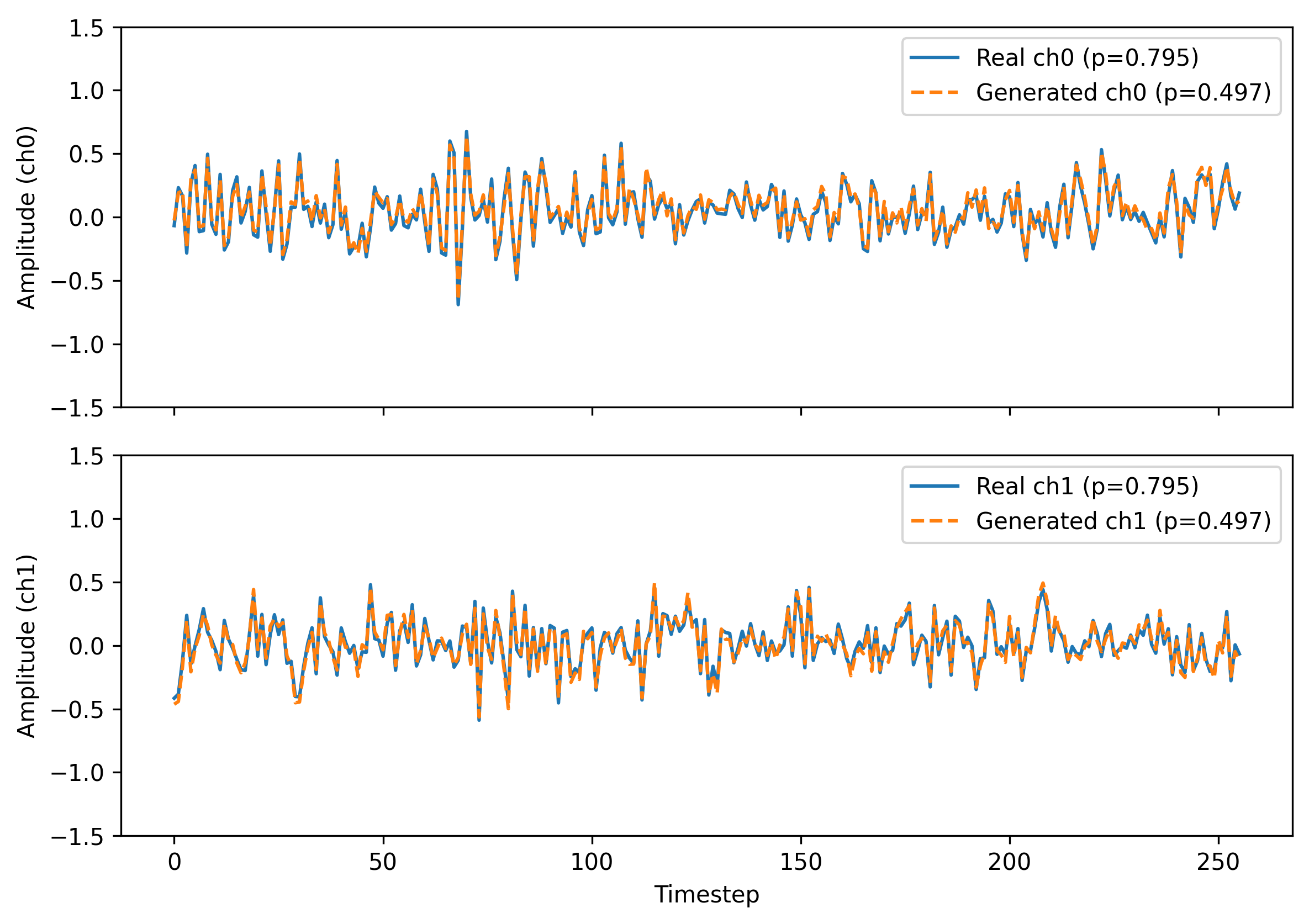}
\caption{Counterfactual generated sample vs. original sample}
\label{fig:fig-cwru-05-cf}
\end{subfigure}
\caption{Target probability 0.5}
\label{fig:fig-cwru-05}
\end{subfigure}

\vspace{0.6em}

\begin{subfigure}[t]{\textwidth}
\centering
\begin{subfigure}[t]{0.49\textwidth}
\centering
\includegraphics[width=\linewidth]{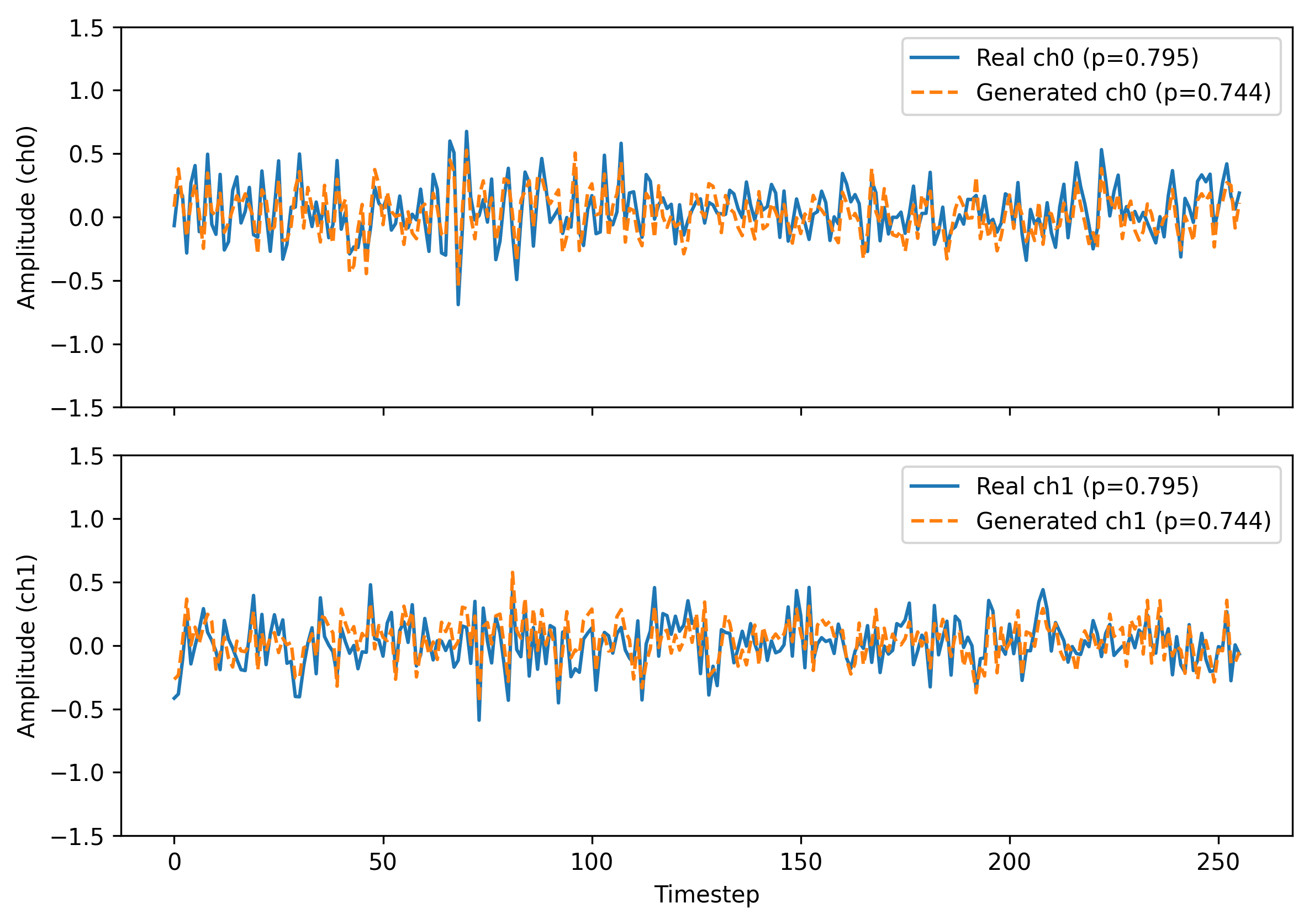}
\caption{PR-GAN generated sample vs. original sample}
\label{fig:fig-cwru-075-gan}
\end{subfigure}
\hfill
\begin{subfigure}[t]{0.49\textwidth}
\centering
\includegraphics[width=\linewidth]{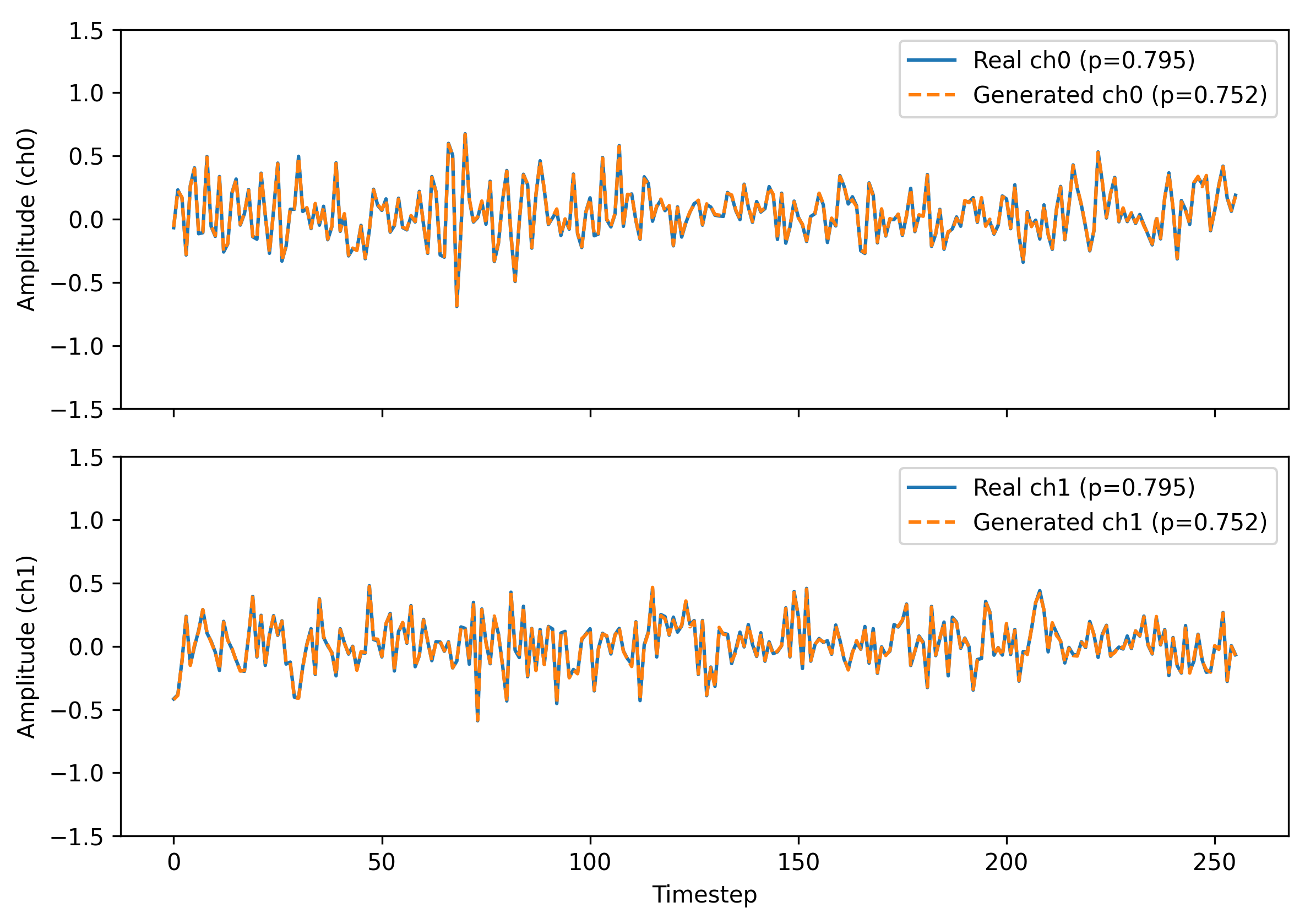}
\caption{Counterfactual generated sample vs. original sample}
\label{fig:fig-cwru-075-cf}
\end{subfigure}
\caption{Target probability 0.75}
\label{fig:fig-cwru-075}
\end{subfigure}

\caption{Generated vs. original samples for the inner race fault class of CWRU dataset. (The real label refers to the label of the original (input) sample.)}
\label{fig:fig-cwru}
\end{figure*}

\begin{figure*}[t]
\centering

\begin{subfigure}[t]{\textwidth}
\centering
\begin{subfigure}[t]{0.49\textwidth}
\centering
\includegraphics[width=\linewidth]{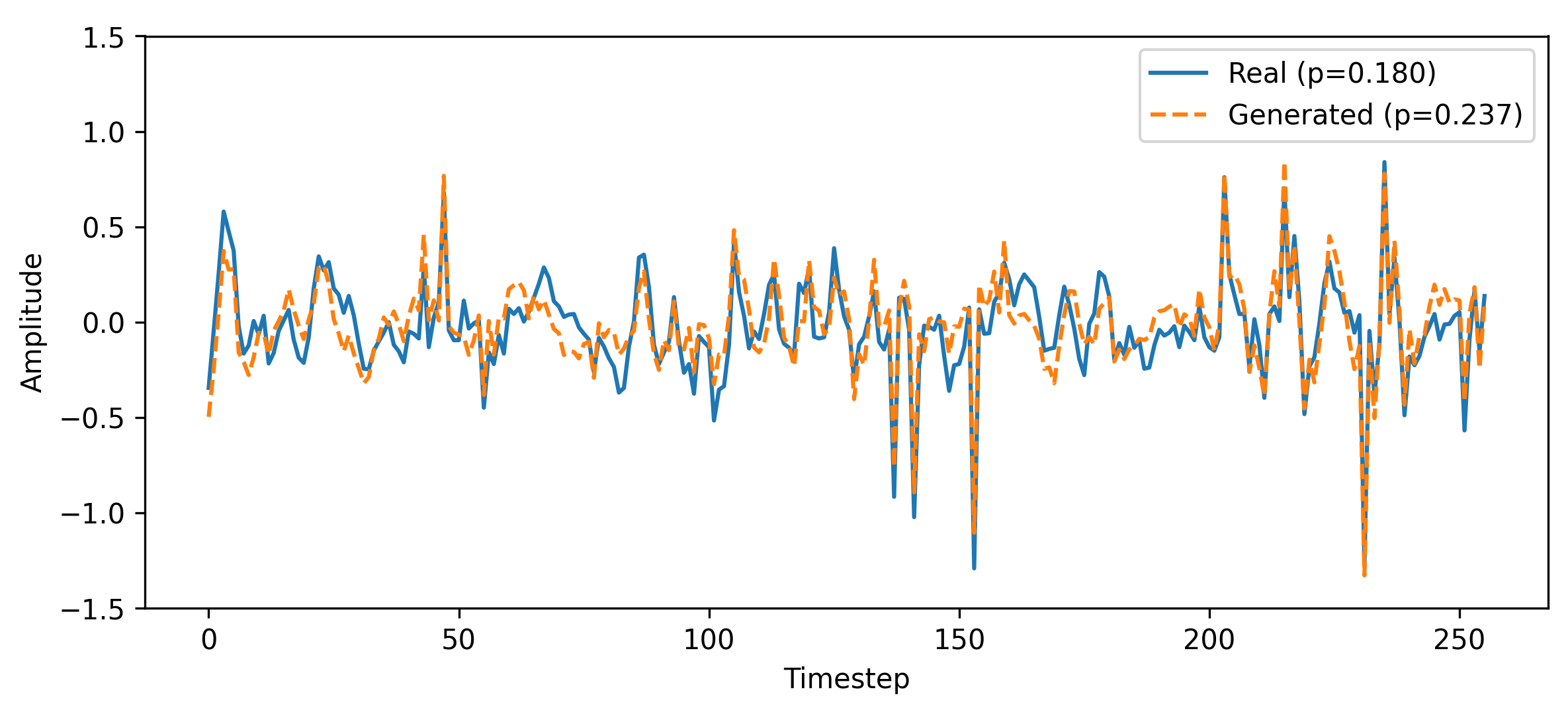}
\caption{PR-GAN generated sample vs. original sample}
\label{fig:fig-paderborn-025-gan}
\end{subfigure}
\hfill
\begin{subfigure}[t]{0.49\textwidth}
\centering
\includegraphics[width=\linewidth]{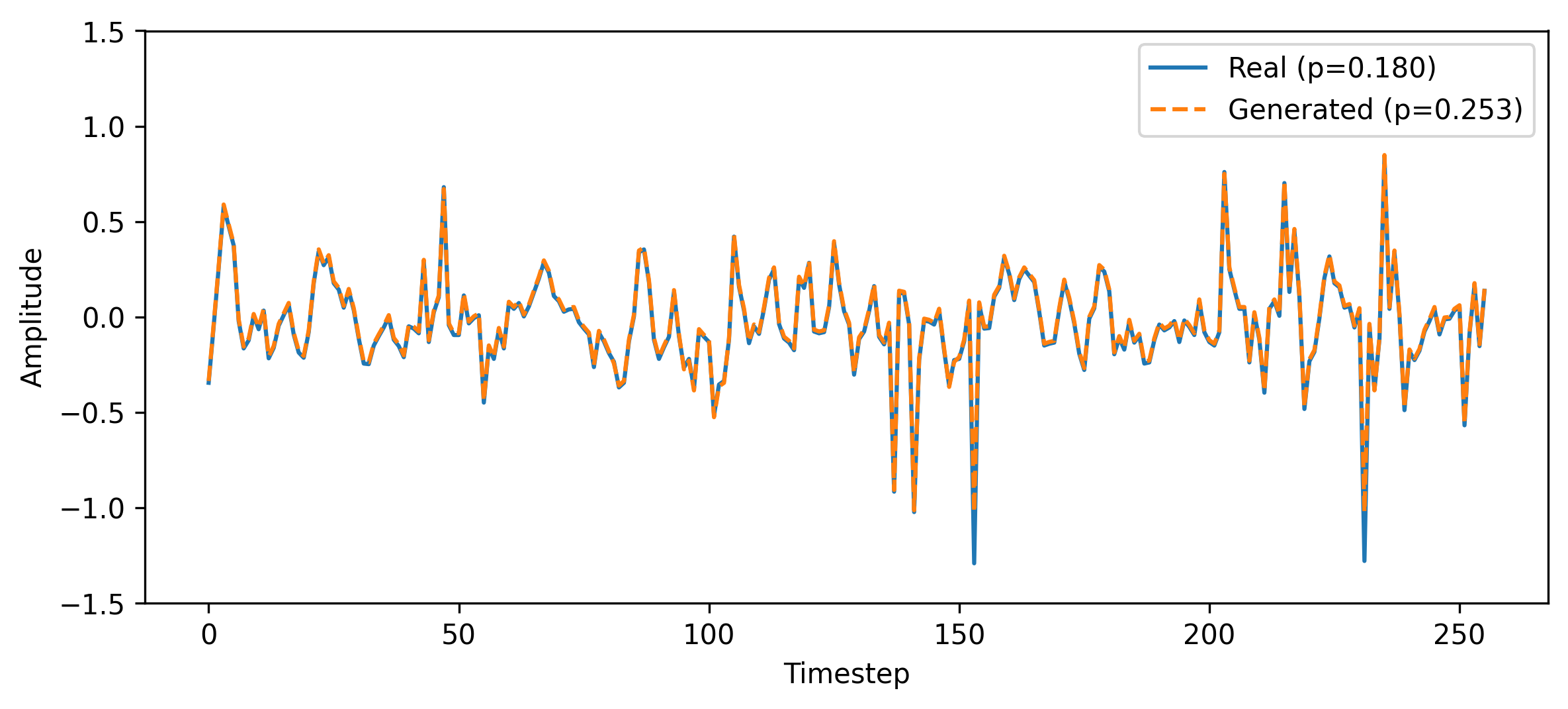}
\caption{Counterfactual generated sample vs. original sample}
\label{fig:fig-paderborn-025-cf}
\end{subfigure}
\caption{Target probability 0.25}
\label{fig:fig-paderborn-025}
\end{subfigure}

\vspace{0.6em}

\begin{subfigure}[t]{\textwidth}
\centering
\begin{subfigure}[t]{0.49\textwidth}
\centering
\includegraphics[width=\linewidth]{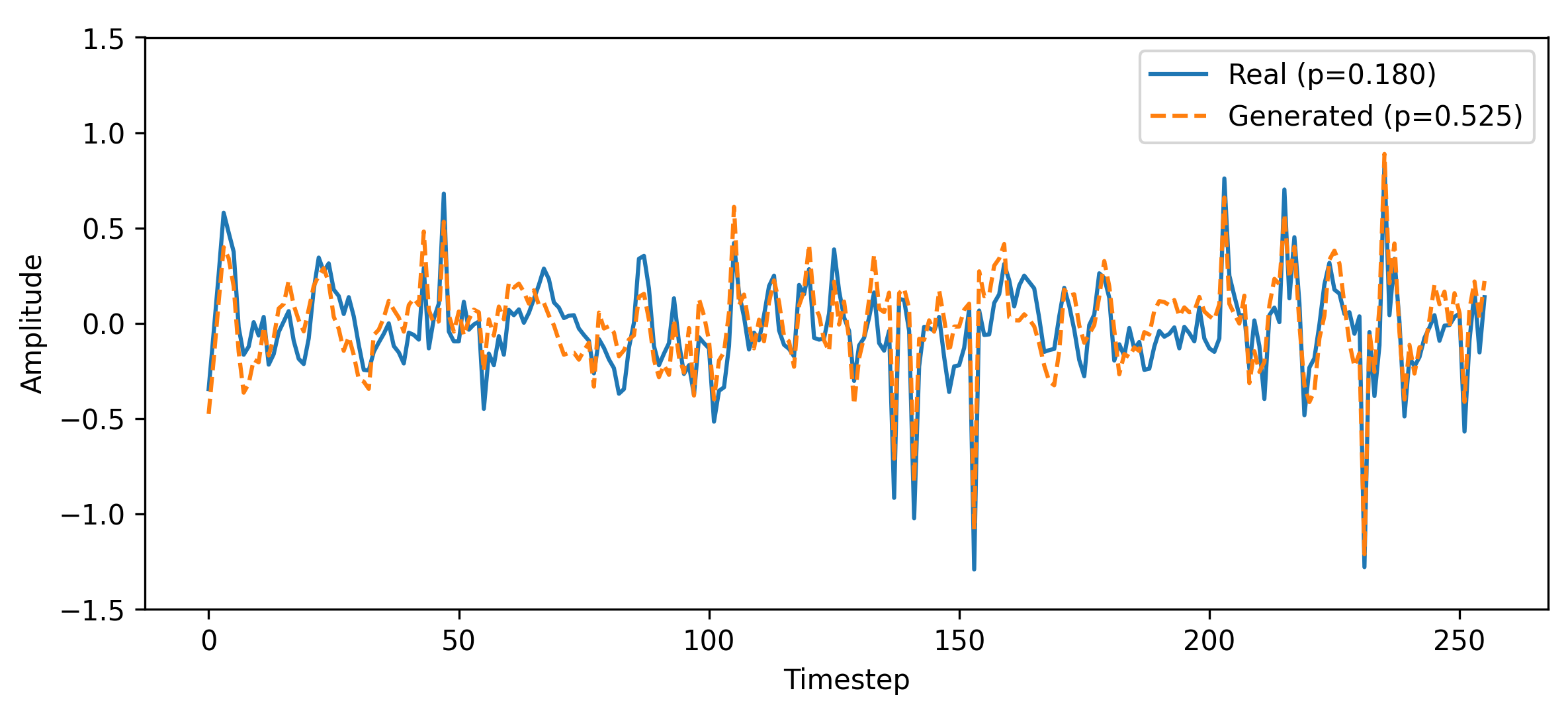}
\caption{PR-GAN generated sample vs. original sample}
\label{fig:fig-paderborn-05-gan}
\end{subfigure}
\hfill
\begin{subfigure}[t]{0.49\textwidth}
\centering
\includegraphics[width=\linewidth]{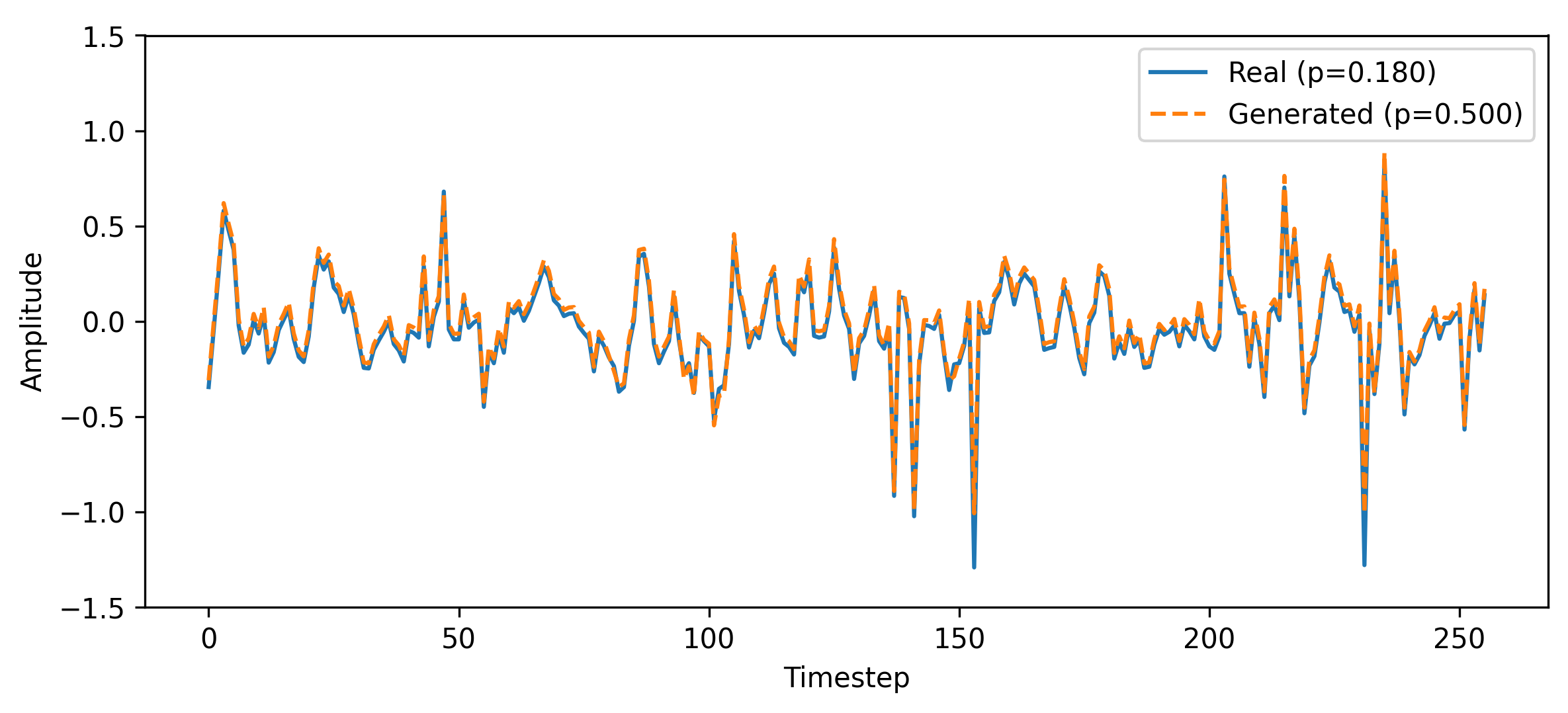}
\caption{Counterfactual generated sample vs. original sample}
\label{fig:fig-paderborn-05-cf}
\end{subfigure}
\caption{Target probability 0.5}
\label{fig:fig-paderborn-05}
\end{subfigure}

\vspace{0.6em}

\begin{subfigure}[t]{\textwidth}
\centering
\begin{subfigure}[t]{0.49\textwidth}
\centering
\includegraphics[width=\linewidth]{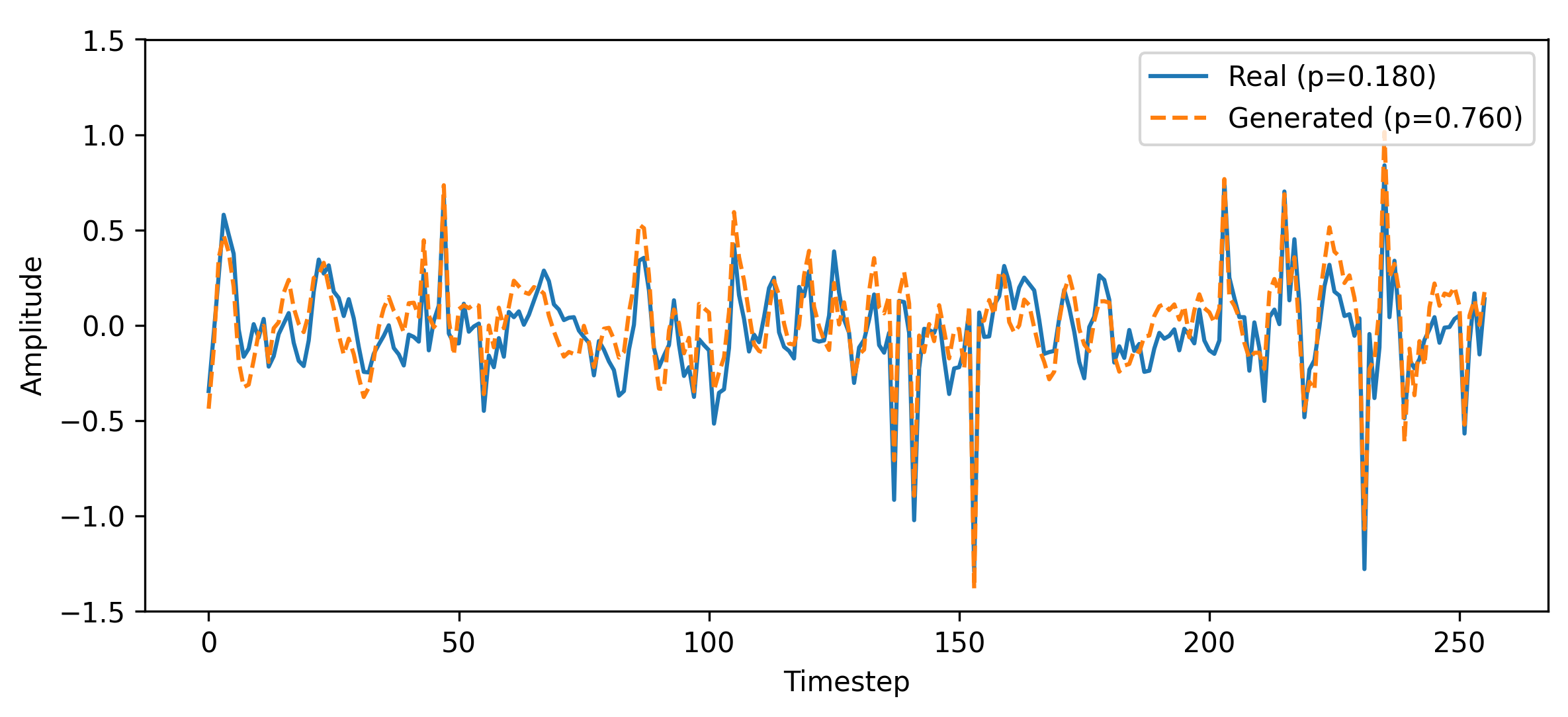}
\caption{PR-GAN generated sample vs. original sample}
\label{fig:fig-paderborn-075-gan}
\end{subfigure}
\hfill
\begin{subfigure}[t]{0.49\textwidth}
\centering
\includegraphics[width=\linewidth]{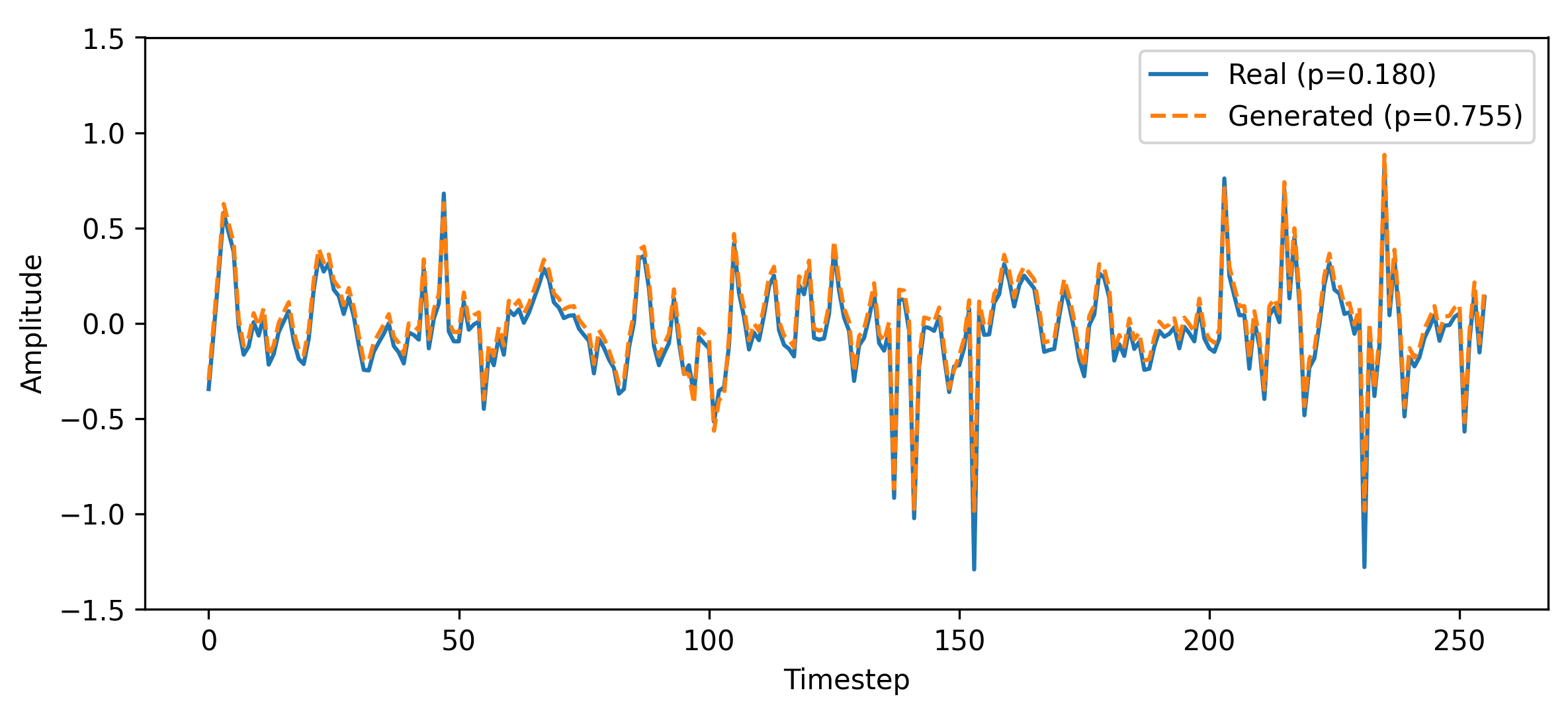}
\caption{Counterfactual generated sample vs. original sample}
\label{fig:fig-paderborn-075-cf}
\end{subfigure}
\caption{Target probability 0.75}
\label{fig:fig-paderborn-075}
\end{subfigure}

\caption{Generated vs. original samples for the real inner ring fault class of Paderborn dataset. (The real label refers to the label of the original (input) sample.)}
\label{fig:fig-paderborn}
\end{figure*}

This section provides additional visual examples showing PR-GAN and CF generated samples with their original source samples for different target probabilities (Figs.~\ref{fig:fig-cwru} and~\ref{fig:fig-paderborn}).

\clearpage

\section*{Acknowledgment}
The first author would like to thank Professor Amir-Hossein Karimi for introducing him to the field of counterfactual and for inspiring his work in this field.

During the preparation of this manuscript, the authors used ChatGPT models solely for language editing, including improvements to grammar, style, and clarity. All scientific content, analysis, and conclusions were developed and verified by the authors.

\balance
\bibliographystyle{cas-model2-names}
\bibliography{biblio}

@article{zhang2020deep,
  author  = {S. Zhang and S. Zhang and B. Wang and T. G. Habetler},
  title   = {Deep Learning Algorithms for Bearing Fault Diagnostics—A Comprehensive Review},
  journal = {IEEE Access},
  volume  = {8},
  pages   = {29857--29881},
  year    = {2020}
}

@article{jalayer2024evaluating,
  author  = {R. Jalayer and M. Jalayer and A. Mor and C. Orsenigo and C. Vercellis},
  title   = {Evaluating Deep Learning Models for Fault Diagnosis of a Rotating Machinery with Epistemic and Aleatoric Uncertainty},
  journal = {arXiv preprint arXiv:2412.18980},
  year    = {2024}
}

@article{wen2018cnn,
  author  = {L. Wen and X. Li and L. Gao and Y. Zhang},
  title   = {A New Convolutional Neural Network-Based Data-Driven Fault Diagnosis Method},
  journal = {IEEE Transactions on Industrial Electronics},
  volume  = {65},
  pages   = {5990--5998},
  year    = {2018}
}

@article{shao2017autoencoder,
  author  = {H. Shao and H. Jiang and H. Zhao and F. Wang},
  title   = {A Novel Deep Autoencoder Feature Learning Method for Rotating Machinery Fault Diagnosis},
  journal = {Mechanical Systems and Signal Processing},
  volume  = {95},
  pages   = {187--204},
  year    = {2017}
}

@article{qian2020imbalance,
  author  = {W. Qian and S. Li},
  title   = {A Novel Class Imbalance-Robust Network for Bearing Fault Diagnosis Utilizing Raw Vibration Signals},
  journal = {Measurement},
  volume  = {156},
  pages   = {107567},
  year    = {2020}
}

@inproceedings{gal2016dropout,
  author    = {Y. Gal and Z. Ghahramani},
  title     = {Dropout as a Bayesian Approximation: Representing Model Uncertainty in Deep Learning},
  booktitle = {Proceedings of the 33rd International Conference on Machine Learning},
  pages     = {1050--1059},
  year      = {2016}
}

@article{luo2021cdcgans,
  author  = {J. Luo and J. Huang and H. Li},
  title   = {A Case Study of Conditional Deep Convolutional Generative Adversarial Networks in Machine Fault Diagnosis},
  journal = {Journal of Intelligent Manufacturing},
  volume  = {32},
  pages   = {407--425},
  year    = {2021}
}

@article{mao2019gan,
  author  = {W. Mao and Y. Liu and L. Ding and Y. Li},
  title   = {Imbalanced Fault Diagnosis of Rolling Bearing Based on Generative Adversarial Network: A Comparative Study},
  journal = {IEEE Access},
  volume  = {7},
  pages   = {9515--9530},
  year    = {2019}
}

@article{tang2021wgan,
  author  = {H. Tang and S. Gao and L. Wang and X. Li and B. Li and S. Pang},
  title   = {A Novel Intelligent Fault Diagnosis Method for Rolling Bearings Based on Wasserstein Generative Adversarial Network and Convolutional Neural Network under Unbalanced Dataset},
  journal = {Sensors},
  volume  = {21},
  pages   = {6754},
  year    = {2021}
}

@article{wachter2018gdpr,
  author  = {S. Wachter and B. Mittelstadt and C. Russell},
  title   = {Counterfactual Explanations without Opening the Black Box: Automated Decisions and the GDPR},
  year    = {2018}
}

@article{liu2025aligned,
  author  = {Shen Liu and Jinglong Chen and Zhen Shi and Liuyang Song and Shuilong He},
  title   = {Representations aligned counterfactual domain learning for open-set fault diagnosis under speed transient conditions},
  journal = {Knowledge-Based Systems},
  year    = {2025}
}

@article{yliu2024fewshot,
  author  = {Yunpeng Liu and Hongkai Jiang and Renhe Yao and Tao Zeng},
  title   = {Counterfactual-augmented few-shot contrastive learning for machinery intelligent fault diagnosis with limited samples},
  journal = {Mechanical Systems and Signal Processing},
  year    = {2024}
}

@article{wang2025cfdata,
  author  = {Chong Wang and Jie Liu and Junwei Cao and Xi Chen and Lihua Chen and Yindong Ji},
  title   = {Counterfactual Data Generation Method for Fault Diagnosis of Complex Electromechanical Systems},
  journal = {IEEE Transactions on Instrumentation and Measurement},
  year    = {2025}
}

@article{ahang2022synthesizing,
  author  = {Maryam Ahang and Masoud Jalayer and Ardeshir Shojaeinasab and Oluwaseyi Ogunfowora and Todd Charter and Homayoun Najjaran},
  title   = {Synthesizing Rolling Bearing Fault Samples in New Conditions: A Framework Based on a Modified {CGAN}},
  journal = {Sensors},
  year    = {2022}
}

@article{cheng2024cvae,
  author  = {Xiaohan Cheng and Yuxin Lu and Zhihao Liang and Lei Zhao and Yuandong Gong and Meng Wang},
  title   = {A Bearing Fault Diagnosis Method in Scenarios of Imbalanced Samples and Insufficient Labeled Samples},
  journal = {Applied Sciences},
  year    = {2024}
}

@article{ruan2023improvement,
  author  = {Diwang Ruan and Xuran Chen and Clemens G{\"u}hmann and Jianping Yan},
  title   = {Improvement of {Generative Adversarial Network} and Its Application in Bearing Fault Diagnosis: A Review},
  journal = {Lubricants},
  year    = {2023}
}

@article{jo2022obgan,
  author  = {Wonkeun Jo and Dongil Kim},
  title   = {OBGAN: Minority oversampling near borderline with generative adversarial networks},
  journal = {Expert Systems with Applications},
  year    = {2022}
}

@article{delaney2021instance,
  author        = {Eoin Delaney and Derek Greene and Mark T. Keane},
  title         = {Instance-based Counterfactual Explanations for Time Series Classification},
  year          = {2021},
  eprint        = {2009.13211},
  archivePrefix = {arXiv}
}

@article{wang2024glacier,
  author  = {Zhendong Wang and Isak Samsten and Ioanna Miliou and Rami Mochaourab and Panagiotis Papapetrou},
  title   = {Glacier: guided locally constrained counterfactual explanations for time series classification},
  journal = {Machine Learning},
  year    = {2024}
}

@article{xu2025zero,
  author  = {Juan Xu and Hui Kong and Xu Ding and Xiaohui Yuan},
  title   = {Counterfactual Inference for Generalized Zero-Shot Compound-Fault Diagnosis},
  journal = {IEEE Transactions on Instrumentation and Measurement},
  year    = {2025}
}

@article{liu2024batch,
  author  = {Zhong Liu and Xuyang Lou},
  title   = {Fault diagnosis based on counterfactual inference for the batch fermentation process},
  journal = {ISA Transactions},
  year    = {2024}
}

@article{gulrajani2017improved,
  author    = {Ishaan Gulrajani and Faruk Ahmed and Martin Arjovsky and Vincent Dumoulin and Aaron Courville},
  title     = {Improved Training of Wasserstein {GAN}s},
  year      = {2017},
  eprint        = {1704.00028},
  archivePrefix = {arXiv}
}

@article{salimans2016improved,
  author    = {Tim Salimans and Ian Goodfellow and Wojciech Zaremba and Vicki Cheung and Alec Radford and Xi Chen},
  title     = {Improved Techniques for Training {GAN}s},
  year      = {2016},
  eprint        = {1606.03498},
  archivePrefix = {arXiv}
}

@article{alain2014dae,
  author  = {Guillaume Alain and Yoshua Bengio},
  title   = {What Regularized Auto-Encoders Learn from the Data-Generating Distribution},
  journal = {Journal of Machine Learning Research},
  year    = {2014}
}

@article{nguyen2017ppgn,
  author    = {Anh Nguyen and Jeff Clune and Yoshua Bengio and Alexey Dosovitskiy and Jason Yosinski},
  title     = {Plug \& Play Generative Networks: Conditional Iterative Generation of Images in Latent Space},
  year      = {2017}
}

@article{Bengio2013-GDAE,
  author       = {Yoshua Bengio and Li Yao and Guillaume Alain and Pascal Vincent},
  title        = {Generalized Denoising Auto-Encoders as Generative Models},
  year         = {2013},
  eprint        = {1305.6663},
  archivePrefix = {arXiv}
}

@article{Getreuer2012-ROF-SB,
  author  = {Pascal Getreuer},
  title   = {Rudin--Osher--Fatemi Total Variation Denoising using Split Bregman},
  journal = {Image Processing On Line},
  year    = {2012}
}

@article{Chhabra-GAP,
  author       = {Sachin Chhabra and Hemanth Venkateswara and Baoxin Li},
  title        = {Generative Alignment of Posterior Probabilities for Source-Free Domain Adaptation}
}

@inproceedings{Guo2017-Calibration,
  author    = {Chuan Guo and Geoff Pleiss and Yu Sun and Kilian Q. Weinberger},
  title     = {On Calibration of Modern Neural Networks},
  booktitle = {Proceedings of the 34th International Conference on Machine Learning},
  year      = {2017}
}

@article{Chen2023-ESLEBS,
  author  = {Chao Chen and Wei Shen and Chenhao Yang and Wei Fan and Xin Liu and Ying Li},
  title   = {A New Safe-Level Enabled Borderline-SMOTE for Condition Recognition of Imbalanced Dataset},
  journal = {IEEE Transactions on Instrumentation and Measurement},
  year    = {2023}
}

@article{Guo2024-SVBorderlineSMOTE,
  author  = {Jiaqi Guo and Haiyan Wu and Xiaolei Chen and Weiguo Lin},
  title   = {Adaptive SV-Borderline SMOTE-SVM algorithm for imbalanced data classification},
  journal = {Applied Soft Computing},
  year    = {2024}
}

@article{Neupane2020BearingFaultCWRUReview,
  title   = {Bearing Fault Detection and Diagnosis Using Case Western Reserve University Dataset With Deep Learning Approaches: A Review},
  author  = {Neupane, Dhiraj and Seok, Jongwon},
  journal = {IEEE Access},
  year    = {2020},
}

@inproceedings{lessmeier2016kat_benchmark,
  title   = {Condition Monitoring of Bearing Damage in Electromechanical Drive Systems by Using Motor Current Signals of Electric Motors: A Benchmark Data Set for Data-Driven Classification},
  author  = {Lessmeier, Christian and Kimotho, James Kuria and Zimmer, Detmar and Sextro, Walter},
  booktitle = {European Conference of the Prognostics and Health Management Society 2016},
  year    = {2016}
}

@misc{BlakeBasicVibrationTheory,
  author = {Blake, Ralph E.},
  title  = {Basic Vibration Theory},
  note   = {Chapter 2}
}

@misc{Gouk2020LipschitzContinuity,
  author        = {Gouk, Henry and Frank, Eibe and Pfahringer, Bernhard and Cree, Michael J.},
  title         = {Regularisation of Neural Networks by Enforcing Lipschitz Continuity},
  year          = {2020},
  eprint        = {1804.04368},
  archivePrefix = {arXiv},
  primaryClass  = {stat.ML}
}

@misc{Ju2017EnsembleCNN,
  author        = {Ju, Cheng and Bibaut, Aurelien and van der Laan, Mark J.},
  title         = {The Relative Performance of Ensemble Methods with Deep Convolutional Neural Networks for Image Classification},
  year          = {2017},
  eprint        = {1704.01664},
  archivePrefix = {arXiv},
  primaryClass  = {stat.ML}
}

@misc{CWRU_BearingDataCenter,
  author       = {{Case Western Reserve University, Case School of Engineering}},
  title        = {Bearing Data Center},
  year         = {n.d.},
  howpublished = {\url{https://engineering.case.edu/bearingdatacenter}},
  note         = {Accessed: 2026-01-07}
}

@misc{PaderbornBDC_DataSetsDownload,
  author       = {{Design and Drive Technology (KAt), Paderborn University}},
  title        = {{Data Sets and Download} --- {Bearing Data Center}},
  howpublished = {Website},
  year         = {n.d.},
  url          = {https://mb.uni-paderborn.de/en/kat/research/bearing-datacenter/data-sets-and-download},
  urldate      = {2026-01-07},
  note         = {Accessed January 7, 2026}
}
\end{document}